\documentclass[lettersize,journal]{IEEEtran}
\usepackage{amsmath,amsfonts}

\usepackage{times}
\usepackage{soul}
\usepackage{url}
\usepackage[hidelinks]{hyperref}
\usepackage[utf8]{inputenc}
\usepackage[small]{caption}
\usepackage{graphicx}
\usepackage{amsmath}
\usepackage{amsthm}
\usepackage{booktabs}
\usepackage{algorithm}
\usepackage{algorithmic}
\usepackage[switch]{lineno}
\usepackage{bm}
\usepackage{color}
\usepackage{dsfont}
\usepackage{amssymb}
\usepackage{pifont}
\usepackage{multirow} 
\usepackage{array}
\usepackage{xcolor}
\usepackage{colortbl}
\usepackage[thicklines]{cancel}

\usepackage{xcolor}
\usepackage{colortbl}
\usepackage{amssymb}

\makeatletter
\def\ps@IEEEtitlepagestyle{
  \def\@oddfoot{\mycopyrightnotice}
  \def\@evenfoot{}
}
\def\mycopyrightnotice{
  {\footnotesize
  \begin{minipage}{\textwidth}
  \centering
  Copyright~\copyright~2024 IEEE. Personal use of this material is permitted. However, permission to use this \\ material for any other purposes must be obtained from the IEEE by sending a request to pubs-permissions@ieee.org.
  \end{minipage}
  }
}

\begin{document}

\def\logouser{\includegraphics[height=10pt]{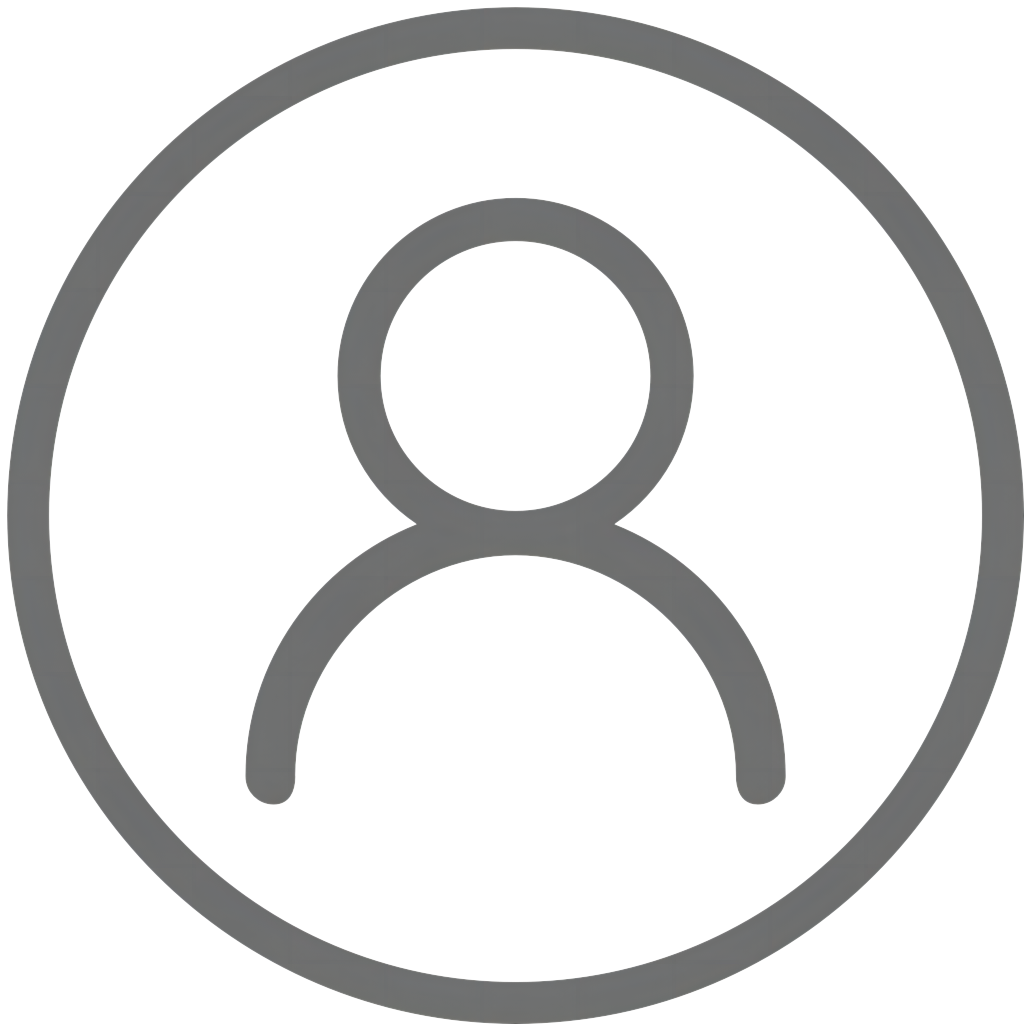}}
\def\logorespomse{\includegraphics[height=10pt]{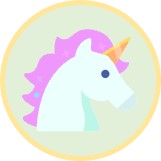}}

\def\logo{\includegraphics[height=23pt]{image/icon_sour.jpg}}

\def\logomini{\includegraphics[height=9pt]{image/icon_sour.jpg}}

\title{GPT4Ego: Unleashing the Potential of Pre-trained Models for Zero-Shot Egocentric Action Recognition}

\author{Guangzhao Dai, Xiangbo Shu, Wenhao Wu, Rui Yan, and Jiachao Zhang


\thanks{
        \textit{G. Dai and X. Shu are with the School of Computer Science and Engineering, Nanjing University of Science and Technology, Nanjing 210094, China E-mail: guangzhaodai@njust.edu.cn; shuxb@njust.edu.cn. Corresponding author: X. Shu.}

       \textit{W. Wu is with the School of Computer Science, The University of Sydney, NSW 2008, Australia. E-mail: wenhao.wu@sydney.edu.au.}
    
       \textit{R. Yan is with the Department of Computer Science and Technology, Nanjing University, Nanjing 210023, China. E-mail: yanrui6019@gmail.com.}

       \textit{J. Zhang is with the Artificial Intelligence Industrial Technology Research Institute, Nanjing Institute of Technology, Nanjing 211167, China. E-mail: zhangjc07@foxmail.com.}
        }
}


\markboth{LATEX}%
{Shell \MakeLowercase{\textit{et al.}}: Bare Demo of IEEEtran.cls for IEEE Journals}


\maketitle
\begin{abstract}
Vision-Language Models (VLMs), pre-trained on large-scale datasets, have shown impressive performance in various visual recognition tasks. This advancement paves the way for notable performance in some egocentric tasks, Zero-Shot Egocentric Action Recognition (ZS-EAR), entailing VLMs zero-shot to recognize actions from first-person videos enriched in more realistic human-environment interactions. Typically, VLMs handle ZS-EAR as a global video-text matching task, which often leads to suboptimal alignment of vision and linguistic knowledge. We propose a refined approach for ZS-EAR using VLMs, emphasizing fine-grained concept-description alignment that capitalizes on the rich semantic and contextual details in egocentric videos. In this work, we introduce a straightforward yet remarkably potent VLM framework, \textit{aka} GPT4Ego, designed to enhance the fine-grained alignment of concept and description between vision and language. 
Specifically, we first propose a new Ego-oriented Text Prompting (EgoTP$\spadesuit$) scheme, which effectively prompts action-related text-contextual semantics by evolving word-level class names to sentence-level contextual descriptions by ChatGPT with well-designed chain-of-thought textual prompts. Moreover, we design a new Ego-oriented Visual Parsing (EgoVP$\clubsuit$) strategy that learns action-related vision-contextual semantics by refining global-level images to part-level contextual concepts with the help of SAM. Extensive experiments demonstrate GPT4Ego significantly outperforms existing VLMs on three large-scale egocentric video benchmarks, \emph{i.e.}, EPIC-KITCHENS-100 (33.2\%{\color{teal}\bm{$\uparrow$}$_{\bm{+9.4}}$}), EGTEA (39.6\%{\color{teal}\bm{$\uparrow$}$_{\bm{+5.5}}$}), and CharadesEgo (31.5\%{\color{teal}\bm{$\uparrow$}$_{\bm{+2.6}}$}). In addition, benefiting from the novel mechanism of fine-grained concept and description alignment, GPT4Ego can sustainably evolve with the advancement of ever-growing pre-trained foundational models. We hope this work can encourage the egocentric community to build more investigation into pre-trained vision-language models.

\end{abstract}


\section{Introduction}

\IEEEPARstart {V}{ision}-Language Models (VLMs), leveraging large-scale contrastive-based image-text pre-training like CLIP~\cite{vanila_clip} and Florence~\cite{Florence}, have shown remarkable zero-shot generalization abilities in video understanding.
Central to this field, egocentric action recognition (EAR) focuses on identifying human actions in first-person videos, gaining increasing attention for its broad applications in areas such as human-object interaction~\cite{mm22hobj}, sports analysis~\cite{wang2021interactive}, and video summarization~\cite{mm22videosum}. 
The strategy of pre-training these versatile VLMs on egocentric videos and adapting them for Zero-Shot Egocentric Action Recognition (ZS-EAR) is emerging as an effective and promising approach.

Recent research has begun to adapt the robust capabilities of {VLMs for {ZS-EAR}, \emph{e.g.}, EgoTV~\cite{egotv}, EgoVLP \cite{egovlp_NeurIPS2022}, EgoVLPv2 \cite{EgoVLPv2_ICCV2023}, and LAVILA \cite{lavila_2023_CVPR}. This performance boost is attributed primarily to two factors: i) the rich semantics provided by large-scale egocentric datasets like Ego4D \cite{ego4d_cvpr2022}; and ii) the video-text matching in the feature space to some extent. However, most VLM-based methods for ZS-EAR directly treat it as coarse-grained global video-text alignment that utilizes natural word-level textual (\emph{e.g.}, the class name) representation and global visual representation (\emph{e.g.}, the class token), resulting in poor semantic alignment between vision and language, as depicted in Figure~\ref{fig2}. Therefore, this leads to a question: \textit{have we fully aligned the vision-language semantic in VLMs for ZS-EAR?}

\begin{figure}[t!]
    \centering
   \includegraphics[width=\linewidth]{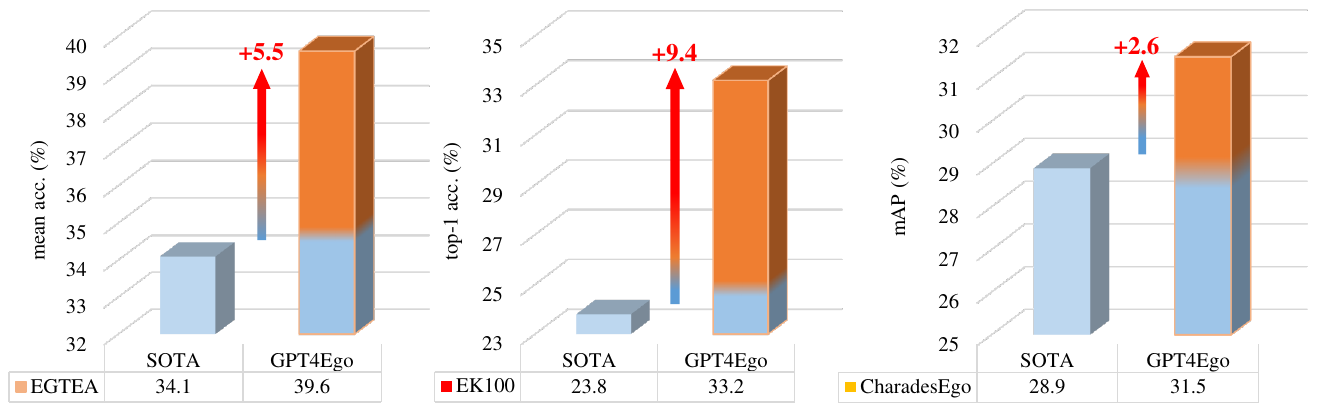}
    \caption{Our \textbf{GPT4Ego} gains significant performance compared with SOTAs in Zero-Shot Egocentric Action Recognition (ZS-EAR) task, by prompting more visual concepts and textual descriptions as the contextual semantics.}
    \label{fig1}
\end{figure}

In this study, the answer is No. In fact, human perceives actions not only based on coarse-grained global video-text alignment but also based on the fine-grained concept-description alignment between vision and language, \textit{i.e.}, action-related diverse textual descriptions and action-related refined visual concepts. Take the action ``\texttt{wash fork}'' as an example, understanding this action requires not only matching global video and text representation, but also the involvement of action-related diverse textual descriptions and refined visual concepts indirectly, \textit{e.g.}, running faucets, sinks, detergents, and rags appeared in the video, which allows us to better understand the human actions in egocentric videos. 

To improve the semantic alignment of vision and language, one naive solution is to utilize larger web-scale video-text pairs to pre-train VLMs. However, such a solution not only additionally requires more cherished video-text pairs than image-text data, but also makes the already expensive VLM training even worse, resulting in poor generalization. The other solution is to generate text descriptions in a zero-shot manner by utilizing an offline text generation model, such as GPT-3 \cite{GPT-3}, GPT-4 \cite{GPT-4}, LLaMA \cite{LLAMA}, and PALM \cite{PALM}. As an outstanding member of GPT family, ChatGPT is a versatile chat robot that exhibits extraordinary zero-shot abilities like language generation and in-context learning, and it also can provide elaborate answers following the user’s instructions. Thus, we turn our attention to such a powerful text generation tool for handling such challenging poor semantic descriptions in egocentric videos. Moreover, recent studies such as GPT4Image \cite{GPT4Image} and GPT4Vis~\cite{GPT4Vis} have excelled in zero-shot image and video perception tasks by generating rich text descriptions with ChatGPT \cite{ren2023chatgpt}. This inspires us to employ ChatGPT for refining initial class names in egocentric actions, generating the corresponding action-related diverse textual descriptions.
                
After elevating text-contextual semantics by generating more action-related textual descriptions in the language domain, a similar requirement can be also brought in the visual representation for fine-grained matching. Commonly, the visual representation in VLMs is usually obtained only from a global perspective by implementing video-text pre-training and then utilizing class tokens to roughly represent video. Intuitively, the more refined visual concepts (\textit{e.g.}, running faucets, sinks, detergents, and rags, depicted in Figure{\color{red}~\ref{fig2}}.) in global video scenes may facilitate fine-grained matching with diverse textual descriptions. One possible solution to label all object's bounding boxes appeared in the videos. However, manually labeling large-scale video datasets is labor- and time-consuming. Another automated solution is to parse visual concepts using zero-shot visual foundational models, such as SAM, which is trained using an extensive dataset of over 1 billion segmentation masks. 
SAM's zero-shot capability in segmenting diverse objects across multiple visual tasks inspires us to utilize it for parsing refined visual contexts in egocentric videos, thus avoiding labor-intensive labeling.

Based on this, we propose a novel VLMs-based paradigm for zero-shot egocentric action recognition, \textit{aka} \textbf{GPT4Ego}: i) improve the fine-grained {concept-description} alignment of vision and language besides a simple global video-text matching; ii) break the technical barrier of low-accuracy performance in zero-shot action recognition tasks in egocentric videos with the help of proposed two insightful designs, \textit{i.e.}, Ego-oriented Text Prompting (EgoTP$\spadesuit$) and  Ego-oriented Visual Parsing (EgoVP$\clubsuit$). Specifically, in EgoTP$\spadesuit$, the word-level class names are evolved to sentence-level contextual descriptions by commanding ChatGPT step-by-step following well-designed chain-of-thought textual prompts, thus effectively prompting action-related text-contextual semantics. Moreover, in EgoVP$\clubsuit$, the global-level images are refined to part-level contextual concepts with the help of SAM, thus effectively prompting action-related vision-contextual semantics. To the best of our knowledge, \textbf{GPT4Ego} is the first work to achieve such two-fold gains in the community of the egocentric action understanding.

In summary, our main contributions are as follows:



\begin{itemize}

    \item We rethink the task of {Zero-Shot Egocentric Action Recognition (ZS-EAR)}, and introduce a simple yet effective VLM framework, named \textbf{GPT4Ego}. To the best of our knowledge, GPT4Ego is the first work that integrates SAM and GPT into VLMs for fine-grained semantic alignment between vision and language by prompting more visual concepts and textual descriptions as the contextual semantics. \par 

    \item {We propose a new Ego-oriented Text Prompting (EgoTP$\spadesuit$) scheme, which effectively prompts action-related text-contextual semantics by evolving word-level class names to sentence-level contextual descriptions by ChatGPT with well-designed chain-of-thought prompts.} \par
    
    \item {We design a new Ego-oriented Visual Parsing (EgoVP$\clubsuit$) strategy, which learns action-related vision-contextual semantics by refining global-level images to part-level contextual concepts with the help of SAM.} \par
    
    \item Extensive experiments conducted on three egocentric video benchmarks demonstrate the state-of-the-art performance achieved by \textbf{GPT4Ego}, namely significant performance on EPIC-KITCHENS-100 (33.2\%{\color{teal}\bm{$\uparrow$}$_{\bm{+9.4}}$}) \cite{damen2020rescaling}, EGTEA (39.6\%{\color{teal}\bm{$\uparrow$}$_{\bm{+5.5}}$}) \cite{egtea}, and CharadesEgo (31.5\%{\color{teal}\bm{$\uparrow$}$_{\bm{+2.6}}$}) \cite{Charades-ego}, respectively.
\end{itemize}

{The remainder of this paper is organized as follows. In Section II, we introduce the related work of ZS-EAR. In Section III, we illustrate the details of the proposed GPT4Ego. In Section IV, we illustrate the experimental results and analysis of the proposed GPT4Ego. Finally, we present the conclusion and discussion of this paper in Section V.}
\IEEEpubidadjcol
\section{Related Works}
{In this section, we first summarize the recent methods related to this paper and then introduce the main characteristics of these works, focusing on three primary directions: Zero-shot Egocentric Action Recognition (Section{\color{red}~\ref{section 3.1}}), Visual-language Learning (Section{\color{red}~\ref{section 3.2}}), and Visual-language Learning (Section{\color{red}~\ref{section 3.3}}).}
\subsection{Zero-shot Egocentric Action Recognition}
\label{section 3.1}
{Compared with exocentric videos~\cite{carreira2017quo,tang2019coherence,shu2015weakly,shu2022multi,shu2019hierarchical,tmm1,tmm2,tmm3} captured from a fixed and distant viewpoint, egocentric videos carry realistic insights (such as human gazes \cite{egtea}) and viewpoint changes caused by frequent human body movements, leading challenges to the common full supervised egocentric action recognition \cite{herzig2022object}, \cite{zhang2022object}, \cite{wu2019long}, \cite{wang2020symbiotic}, \cite{wang2020symbiotic2}, \cite{ma2022hand}, \cite{escorcia2022sos}.} Vision-Language Models (VLMs) pre-trained on large-scale datasets have achieved impressive performance in various visual recognition tasks. Recently, some researchers have been trying to focus on transferring knowledge from the powerful VLMs and expanding it to {{Zero-Shot Egocentric Action Recognition (ZS-EAR)}} \cite{egotv}, \cite{egovlp_NeurIPS2022}, \cite{EgoVLPv2_ICCV2023}, \cite{lavila_2023_CVPR}, which manifests outstanding performance. Among them, Rishi \textit{et al.}~\cite{egotv} propose Egocentric Task Verification (EgoTV), a benchmark and a synthetic dataset that contains pairs of videos and their task descriptions, which verifies the execution of tasks from egocentric videos based on the natural language description of these tasks. Lin \textit{et al.}~\cite{egovlp_NeurIPS2022} fully explore the recently released Ego4D dataset and create a new egocentric dataset called EgoClip, which collects a large variety of human daily activities and comprises 3.8M clip-text pairs well-chosen from Ego4D. Based on EgoClip, it introduces a pre-training video-text contrastive learning objective called EgoNCE and a development benchmark called EgoMCQ, advancing the pace of ZS-EAR. Recently, Zhao \textit{et al.} propose LAVILA \cite{lavila_2023_CVPR}, an approach to learning video-language representations by leveraging Large Language Models (GPT-2) \cite{GPT-2}, which first densely narrate long videos by LLM, and then uses those narrations to pre-train video models.
For the downstream task, \textit{e.g.,} zero-shot egocentric action recognition, it directly calculates the cosine similarity of the feature of the class name and class token to perform global video-text matching. Although achieve outstanding performance, by constructing another dataset for pre-training or directly expanding video narrations, the above VLMs-based methods for ZS-EAR consider the coarse-grained video-text alignment by utilizing naive word-level textual representation (\textit{e.g.}, class name) and global visual representation (\textit{e.g.}, class token), resulting in sub-optimal alignment between vision and language. 
In this paper, we rethink the task of ZS-EAR and propose an embarrassingly simple yet surprisingly effective VLM framework, named GPT4Ego, which believes that a more graceful paradigm of ZS-EAR should be a fine-grained {concept-semantic} alignment between vision and language in VLMs.

\subsection{Visual-language Learning}
\label{section 3.2}
Recently, Vision-Language Models (VLMs) like CLIP~\cite{vanila_clip} leveraging image-text contrastive learning, have shown impressive performance in a variety of image tasks, \textit{e.g.}, image classification \cite{vln_imgcls}, object detection \cite{vln_objectdet}, image segmentation \cite{vln_img_seg}, and \textit{et al}. Consequently, it also lights researchers in video {tasks~\cite{tmm4,tmm5,tmm6}} that pre-train on video-related datasets, advancing the amount of video-specific VLMS {\cite{text4vis}, \cite{text4vis_ijcv}, \cite{actionclip}, \cite{videoclip}, \cite{xclip}, \cite{GPT4Vis}, \cite{st-adapter}, \cite{aim}, \cite{cap4video}}. VideoCLIP \cite{videoclip} utilizes video-text pairs to pre-train VLM for the video action recognition task in a from-scratch manner, which also suffers from the problems for data-hungry and computationally expensive. ActionCLIP \cite{actionclip} proposes an alternative pipeline, \textit{i.e.}, ``pre-train, prompt and finetune", which alleviates these problems by fine-tuning upon the available image-text pre-trained model. X-CLIP \cite{xclip} introduces a cross-frame attention mechanism and video-specific prompting scheme, which are used to capture the temporal relation of frames and generate the text prompts by utilizing the video content embedding, respectively. Text4vis \cite{text4vis} revisits the classifier with pre-trained models (\textit{e.g.}, CLIP) for improving efficient action recognition. In addition, some works have also focused on the training efficiency of VLM. By freezing the visual encoder and adding some lightweight temporal and spatial adapters, ST-Adapter \cite{st-adapter} and AIM \cite{aim} lower the tuneable parameters and promote efficient video recognition. While these methods show appropriate zero-shot generalizability and effectiveness, all of them simply treat video action recognition as a global-level video-text matching and neglect fine-grained semantics that are essential for egocentric video understanding in the visual and textual domain, which may not be available for ZS-EAR. In this paper, we first rethink ZS-EAR and then present a novel paradigm, \textit{aka} GPT4Ego, to improve the fine-grained semantic alignment of the vision and language in VLMs.

\subsection{Large Pre-trained Models}
\label{section 3.3}
In recent years, {Large Language Models (LLMs)}, such as GPT-3 \cite{GPT-3}, GPT-4 \cite{GPT-4}, LLaMA \cite{LLAMA}, and PALM \cite{PALM}, have manifested strong generated ability owing to the success of large-scale vision-language pre-training in {Natural Language Processing (NLP)}. 
ChatGPT, launched in November 2022 by OpenAI, is a conversational chatbot based on the design of GPT (Generative Pre-trained Transformer) \cite{GPT-2} architecture, designed to generate human-like responses across a wide range of topics and contexts. Later, it also updates some variants, such as the variant GPT-4 \cite{GPT-4} with better contextual understanding and GPT-4V \cite{gpt4v} with image understanding dialog. The above Large models also inspire {Computer Vision (CV)} community, including video understanding~\cite{lin2023mm}, image context reasoning~\cite{liu2023hallusionbench}, optical character recognition (OCR)~\cite{shi2023exploring}, recommender system~\cite{zhou2023exploring}, medical imaging analysis~\cite{yang2023performance,li2023comprehensive}, mathematical logic~\cite{lu2023mathvista} and autonomous driving~\cite{wen2023road}, social media analysis~\cite{lyu2023gpt}, etc. Additionally, the {CV} community is witnessing the emergence of impressive foundational models \cite{blip-2,zhu2023minigpt, Alayrac2022Flamingo,liu2023llava,wang2022internvideo}, like the Segment Anything Model (SAM) \cite{SAM}, which can segment any object by leveraging extensive pre-training on a vast dataset containing over 1 billion segmentation masks. The above works inspire us to utilize the powerful capability of these large pre-trained models, to address the problem of ZS-EAR, i.e., the pretext of the coarse-grained semantic alignment in VLM. To this end, we propose GPT4Ego, which integrates SAM and GPT into VLMs to promote the fine-grained alignment between vision and language. To the best of our knowledge, GPT4Ego is the first work that integrates SAM and GPT into VLMs for promoting semantic alignment between vision and language for ZS-EAR.
\begin{figure*}[t!]
    \centering
    \includegraphics[width=\linewidth]{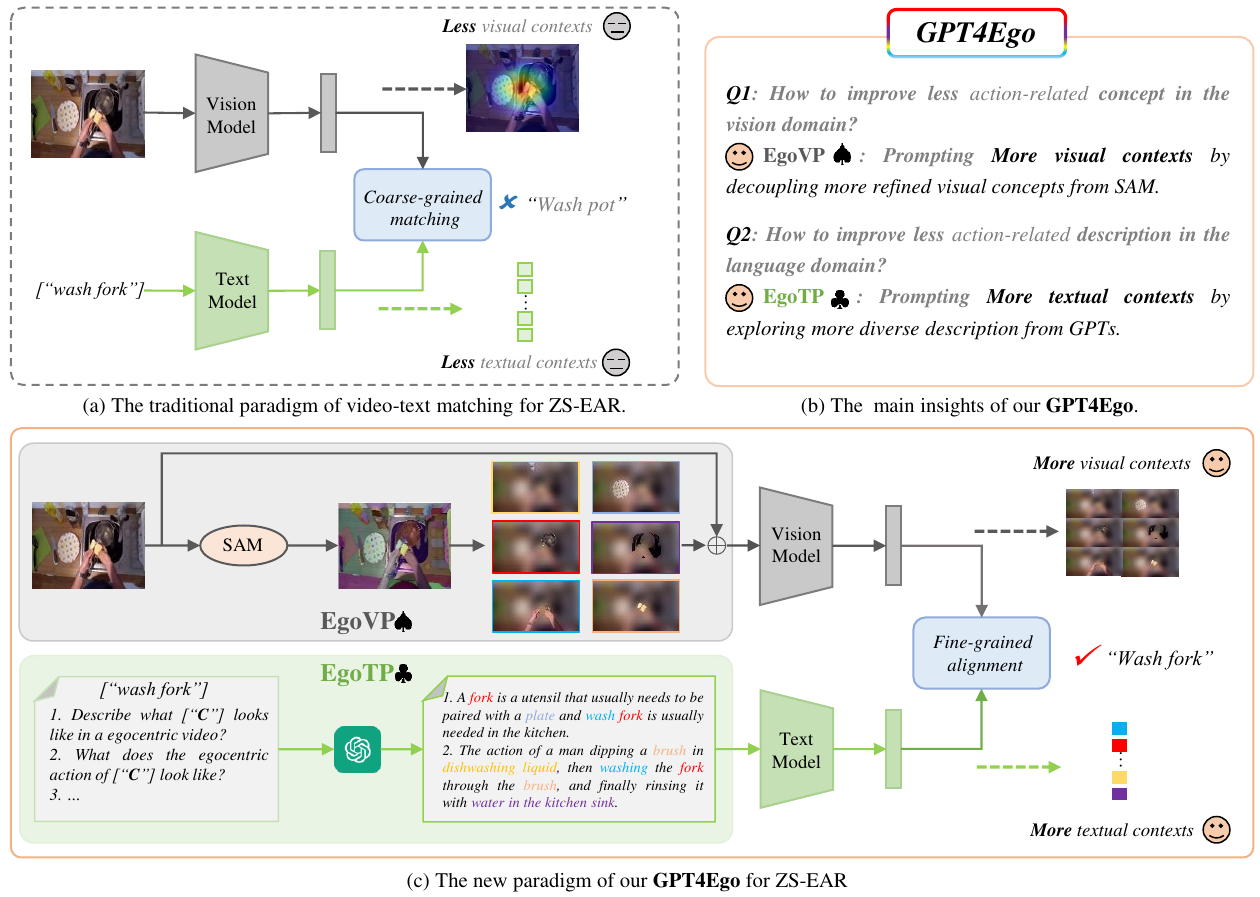}
    \caption{Illustration of pre-trained vision-language models (VLMs) for zero-shot egocentric action recognition (ZS-EAR). (a) The previous VLMs treat ZS-EAR as a coarse-grained global video-text matching task, resulting in poor semantic alignment. (b) The main insights of our proposed \textbf{GPT4Ego} are to answer the limitations (i.e., almost treat ZS-EAR as a coarse-grained global video-text matching task) of previous works, in a two-fold way, i.e., prompting more action-related textual descriptions and prompting more action-related visual concepts by using open resources large language/vision foundation models (e.g., ChatGPT and SAM). (c) The new paradigm of our \textbf{GPT4Ego} after rethinking the task of ZS-EAR, which integrates SAM and GPT into VLMs for promoting the fine-grained semantic alignment between vision and language for ZS-EAR .}
    \label{fig2}
\end{figure*}
\section{Method}\label{sec:three}
{{The proposed framework, named GPT4Ego, to unleashes the potential of pre-trained models for ZS-EAR. It mainly derives from two elaborated designs, including Ego-oriented Text Prompting (EgoTP$\spadesuit$) and Ego-oriented Visual Parsing (EgoVP$\clubsuit$).} In this section, we will first provide the preliminary formulation for the task of ZS-EAR in Section~\ref{Preliminary}. Then, we will give the detail of EgoTP$\spadesuit$ in Section~\ref{EgoTP}. The detail of EgoVP$\clubsuit$ is presented in Section~\ref{EgoVP}. Finally, we will illustrate the objective function in Section~\ref{Objective Function}}   
\subsection{Preliminary: VLM-based ZS-EAR}
\label{Preliminary}
We first describe the pipeline of traditional zero-shot egocentric action recognition (ZS-EAR) based on pre-trained vision-language models (VLMs) and then present our well-designed \textbf{GPT4Ego}. 
Suppose an input egocentric video ${\bm{X}} \in {\mathds{R}}^{T \times 3 \times H \times W}$ consisting of $T$ RGB frames of resolution $H \times W$ and the label of class name $Y$, VLM for ZS-EAR aims to associate the text representation ${\bm{\mathcal{T}_{cls}}}$ (\textit{e.g}., class name) with video representation (\textit{e.g.}, class token) ${\bm{\mathcal{V}_{cls}}}$ directly to perform global video-text matching, as follows
\begin{equation}
    \bm{\mathcal{V}_{cls}} \longleftarrow \mathcal{M}\left ( {\bm{X,T}} \right ) \longrightarrow  {\bm{\mathcal{T}_{cls}}},
    \label{previous_pipline}
\end{equation}
where $\mathcal{M}\left ( \cdot  \right )$  denotes one classical VLM pre-trained on large-scale egocentric video-text datasets, \textit{e.g.}, EgoTV~\cite{egotv}, EgoVLP \cite{egovlp_NeurIPS2022}, EgoVLPv2 \cite{EgoVLPv2_ICCV2023}, LAVILA \cite{lavila_2023_CVPR}), \textit{et al}. However, the VLMs-based paradigm of global video-text matching in these methods causes poor semantic alignment between vision and language due to neglecting the fact, \textit{i.e.}, egocentric videos are more focused on fine-grained semantics of text- and vision-contextual, as depicted in Figure~{\color{red}\ref{fig2}}.

In this paper, we rethink ZS-EAR and present a novel paradigm, \textit{aka} \textbf{GPT4Ego}, to improve the fine-grained semantic alignment of the vision and language in VLMs, including action-related text-contextual semantics and action-related vision-contextual semantics. Here, Eq.~\eqref{previous_pipline} can be reformulated as

\begin{equation}
    \bm{\mathcal{V}_{con}}  \xleftarrow{\rm EgoVP\clubsuit}  {\mathcal{M}\left ( {\bm{X,T}} \right )} \xrightarrow{\rm EgoTP\spadesuit} {\bm{\mathcal{T}_{des}}}.
    \label{new_pipline}
\end{equation}

\textbf{GPT4Ego} mainly contains two new components: \textbf{i) Ego-oriented Text Prompting} (EgoTP$\spadesuit$) that effectively prompt action-related text-contextual semantics by evolving word-level class names to sentence-level contextual descriptions by ChatGPT with well-designed {chain-of-thought} textual prompts; and \textbf{ii) Ego-oriented Visual Parsing} (EgoVP$\clubsuit$) that learns action-related vision-contextual semantics by refining global-level images to part-level contextual concepts with the help of SAM. Specifically, in EgoTP{$\spadesuit$}, we command ChatGPT to process initial class name $Y$ by utilizing several well-designed {chain-of-thought} text prompting instructions, obtaining semantic-diversity text representation $\bm{\mathcal{T}_{des}}$, see Section{{\color{red}~\ref{EgoTP}}} for details. In EgoVP$\clubsuit$, with the help of the capability of SAM that can segment arbitrary objects, we process the original video frames ${\bm{X}}$ to obtain several pixel-level masks of visual concepts, which are then further evolved to refined vision-contextual concepts $\bm{\mathcal{V}_{con}}$ by a scheme of Concept Augmentation, see Section{{\color{red}~\ref{EgoVP}}} for details. 

\subsection{Ego-oriented Text Prompting (EgoTP$\spadesuit$)}
\label{EgoTP}
To improve action-related text-contextual semantics, we present a novel EgoTP$\spadesuit$ scheme, which processes the initial class name $Y$ to more diverse textual descriptions ${\bm{\mathcal{T}_{cls}}}$ by prompting ChatGPT, as depicted in Figure{\color{red}~\ref{fig2}}. 
Inspired by the design of chain-of-thought~\cite{GPT-3} in NLP and the cognitive process of our human visual system, we propose the following approach for setting up text prompts: \bm{\ding{172}} \textbf{Scene description}, \textit{i.e.,} Start by simplifying the given keywords into a scene context; \bm{\ding{173}} \textbf{Question Answering}, \textit{i.e.,} Further enhance the scene details using QA characteristics; \bm{\ding{174}} \textbf{Action Associating}, \textit{i.e.,} Utilize analogies and associations to infer attributes; \bm{\ding{175}} \textbf{Caption Generation}, \textit{i.e.,} Finally generate a concise yet contextual understanding.

Take the action of ``\texttt{wash fork}" as an example,  we enquire ChatGPT with these prompts step by step as follows.
\\ \bm{\ding{172}} \textbf{Scene description}. Given a descriptive scene prompt, ChatGPT can summarize scene by understanding the egocentric domain-specific action, \textit{e.g.}, Input: ``{\color{gray!120} \textit{Describe the egocentric action of [``wash fork"] in detail.}}''; ChatGPT: ``{\color{gray!120}\textit{{Washing fork is a common action in the kitchen and usually involves the following steps: First, dipping a brush in dishwashing liquid, then washing the fork through the brush, and finally rinsing it with water in the kitchen sink}}}''.
\\ \bm{\ding{173}} \textbf{Question Answering.} It can answer the question using a textual prompt, \textit{e.g.}, Input: ``{\color{gray!120}\textit{How can you identify the action of [``wash fork"] in an egocentric video.}}''; ChatGPT: ``{\color{gray!120}\textit{Design and Shape:  A fork is a utensil that is usually paired with a plate. A wash fork typically has a long, thin handle with narrow, pointed tines that are evenly spaced out.}}''.
\\ \bm{\ding{174}} \textbf{Action Associating.} ChatGPT can create associations from keywords using its knowledge base, \textit{e.g.}, Input: ``{\color{gray!120}\textit{What does the egocentric action of [``wash fork"] look like?}}; ChatGPT: ``{\color{gray!120}\textit{The action of washing a fork typically involves holding the fork by the handle and using a cloth or sponge with soap and water to scrub and clean both the tines and the handle.}}".   
\\ \bm{\ding{175}} \textbf{Caption Generation}. For an initial class name, ChatGPT is expected to generate a descriptive caption, \textit{e.g.}, Input: ``{\color{gray!120}\textit{A caption of an egocentric action of [``wash fork"]:}}"; ChatGPT: ``{\color{gray!120}\textit{The person grabs a rag or cloth and holds it in one hand. Then, they pick up the fork in the other hand and proceed to wipe the tines and handle of the fork with the rag, using a back-and-forth motion.}}". 

Based on these {chain-of-thought} text prompts, we traverse all class names in the dataset with \textit{K-th} categories and replace the initial class names of the corresponding positions with the generated diverse textual descriptions by ChatGPT. Finally, the integrated descriptions of each class name are fed into the textual encoder of VLMs, obtaining text-contextual representation ${\bm{\mathcal{T}_{cls}}} \in \mathbb{R}^{K \times D}$.

\subsection{Ego-oriented Visual Parsing (EgoVP$\clubsuit$)}
\label{EgoVP}
In Section{\color{red}~\ref{EgoTP}}, we elevate text semantics by generating diverse textual descriptions, which also brings a similar requirement to the visual representation for better matching. However, the visual representation in typical VLMs is usually given only a macroscopic global signal and cannot focus on local concepts well, owing to the paradigm of common video-text pre-training. Intuitively, more refined visual concepts (\textit{e.g.}, running faucets, sinks, detergents, and rags, depicted in Figure{\color{red}~\ref{fig2}}.) in global video scenes may facilitate fine-grained matching with diverse textual descriptions. To this end, we present an Ego-oriented Visual Parsing strategy (EgoVP$\clubsuit$) to fill this gap (\textit{i.e.}, the pretext of the coarse-grained semantic alignment in VLM.) in two steps, namely Scene Parsing and Concepts Augmentation.
\\ \textbf{Scene Parsing.}
Rather than relying on any labor-intensive object annotation, we turn our attention to automatic scene parsing tools to obtain refined visual concepts in videos. Here, the tool of scene parsing is SAM, an excellent zero-shot segmentation model, whose capability of segmenting anything has been proven in many visual works. Thus, based on SAM, we can obtain pixel-level position masks of all refined concepts in video frames. Specifically, give a frame $\bm x$ in input video clip $\bm X$ with $T$ frames, all masks represented refined visual concepts in $\bm x$ can be calculated as:
\begin{align}
    {\bm M} &= \mathcal{SAM}(\bm x), 
\end{align}
where $\bm M \in \mathbb{R}^{N \times 3 \times H \times W}$ consisting of $N$ pixel-level masks. Note that the number of $N$ is different for each image because SAM is adaptively segmented according to the semantical quantity in each image. Therefore, we set the hyper-parameter $Q (Q<N)$ to make a quantitative selection from the $N$ concepts to obtain the most optimal performance, which will be discussed in detail in Table{\color{red}~\ref{cocepts_table}}. Hence, $\bm M$ is be re-represented as $\bm M^{*} \in \mathbb{R}^{Q \times 3 \times H \times W}$ for the sake of calculability.    
\\ \textbf{Concept Augmentation.} After obtaining the masks of refined visual concepts, we next need to generate corresponding refined visual concepts in that frame. One possible solution is to project these masks with the initial frame directly to generate several mask-based images. However, such naive images only preserve mask-based local information, neglecting contextual information, which is essential for recognizing egocentric actions and has been discussed in Section{\color{red}~\ref{Preliminary}}. More recently, the masking strategy introduced by FGVP \cite{FGVP} projects foreground masks while preserving a certain degree of background information for improving image classification. Motivated by this, we present a Concept Augmentation scheme to generate refined vision-contextual concepts, which can accurately highlight the refined visual concept, reduce background interference, and retain contextual information. Given masks of refined visual concepts in a frame, the augmented image $\bm x_c$ ({\textit{i.e.}, vision-contextual concepts}) can be formulated as:     
\begin{align}
    \bm x_c &= \mathcal{CA} (\bm{x} \odot {\bm M^{*}}),
\end{align} 
{where $\mathcal{CA (\cdot)}$ denotes the operation of Concept Augmentation},  $\bm{x_c} \in \mathbb{R}^{N \times 3 \times H \times W}$ denotes the augmented image with same size of $\bm M$, and $\odot$ denotes an element-wise dot product. Following the above process, for the input video $\bm X$ within $T$ frames, we generate $T$ numbers augmented images as $\bm{X_c} \in \mathbb{R}^{T \times Q \times 3 \times H \times W}$. Finally, we feed the $\bm{X_c}$ into the visual encoder of VLM, to produce a vision-contextual representation $\bm{\mathcal{V}_{con}} \in \mathbb{R}^{1 \times D}$. 

\subsection{Objective Function}
\label{Objective Function}
After obtaining the visual representation $\bm{\mathcal{V}_{con}} \in \mathbb{R}^{1 \times D}$ and text representation ${\bm{\mathcal{T}_{cls}}} \in \mathbb{R}^{K \times D}$, the logits are can be obtained by calculating cosine similarity between them. In addition, the representation of $\bm{\mathcal{V}_{con}} \in \mathbb{R}^{1 \times D}$ and ${\bm{\mathcal{T}_{des}}} \in \mathbb{R}^{K \times D}$ in Eq.~\ref{previous_pipline} is also used for obtaining the residual logits. Thus, the final logits of ZS-EAR can be formulated as:
\begin{equation}
    Logits = Sim({\bm{\mathcal{V}_{con}}} \cdot {{\bm{\mathcal{T}_{des}}}^{T}}) + Sim(\bm{\mathcal{V}_{cls}} \cdot {\bm{\mathcal{T}_{cls}}}^{T}),
\label{objection_function}
\end{equation}
where $Sim(\cdot)$ denotes the function of cosine similarity.

\section{Experiments}\label{sec:four}
We conducted extensive quantitative and qualitative experiments on three public egocentric benchmarks, including EPIC-KITCHENS-100~\cite{damen2020rescaling}, EGTEA~\cite{egtea}, and EGTEA~\cite{egtea}, to evaluate the effectiveness of the proposed method. In this section, we first illustrate the dataset and metrics in Section~\ref{Datasets and Metrics}. Then, we will compare with state-of-the-art methods to illustrate the performance of GPT4Ego in Section~\ref{compare sota}. The implementation details, including Encoder, EgoTP$\spadesuit$, and EgoVP$\clubsuit$, are presented in Section~\ref{Implementation Details}. After that, we perform a set of ablation studies in Section~\ref{Ablation Studies}, to discuss the effect of each component, each text prompt, and the number of visual concepts. Lastly, the qualitative analysis is demonstrated in Section~\ref{Qualitative Analysis}.

\subsection{Datasets and Metrics}
\label{Datasets and Metrics}
We conducted experiments on three classic egocentric video datasets, including EPIC-KITCHENS-100 (EK100)~\cite{damen2020rescaling}, EGTEA~\cite{egtea}, and CharadesEgo~\cite{Charades-ego}. Since our method requires no training, we directly evaluated it on the validation or test sets of these datasets following the comparative methods. The details of the datasets are as follows.

\begin{itemize}
    \item EPIC-KITCHENS-100~\cite{damen2020rescaling} is one of the most popular large-scale egocentric datasets with 89,977 action clips collected over 100 hours in 45 kitchen environments using the head-mounted cameras. It collects the participant's variety of daily activities in the kitchen with a first-person view. For the dataset splitting, the training/validation/testing sets contain 67,217/9,668/13,092 video clips, respectively. Following the previous protocol~\cite{wang2021interactive,wang2020symbiotic2}, we report top-1 and top-5 accuracy of action on the validation set. \par
    \item EGTEA~\cite{egtea} is a large-scale egocentric dataset with 28 hours of daily kitchen activity, which is collected using gaze tracking. It consists of 10,321 video clips and is labeled with 51 noun classes, 19 verb classes, and 106 action classes. For the dataset splitting, the training/testing sets contain 8,299/2,022 video clips, respectively. Following the previous protocol~\cite{sudhakaran2018attention,sudhakaran2019lsta}, we report top-1 accuracy and mean accuracy (mean acc.) on the testing split. \par
    \item CharadesEgo~\cite{Charades-ego} is also a large-scale dataset containing 7,860 videos of daily indoor activities from both first- and third-person views. The dataset annotates 68,536 instances of fine-grained actions from 157 classes. Here, we only use the first-person subset comprising 846 videos for zero-shot testing. We use mean Average Precision (mAP) as the evaluation metric following the previous protocol~\cite{egovlp_NeurIPS2022,lavila_2023_CVPR}.
\end{itemize}

\subsection{Implementation Details}
\label{Implementation Details}
The implementation of the overall framework is carried out on PyTorch \cite{pytorch} in a Linux environment with an NVIDIA GeForce 3090. Based on the architecture of the overall framework, the implementation details of this work mainly can be described in three aspects, including Encoder, EgoTP$\spadesuit$, and EgoVP$\clubsuit$.

\textbf{Encoder}: We use standard LAVILA \cite{lavila_2023_CVPR} with visual-textual architecture as the baseline for all ablation experiments. Among them, the standard TimeSformer (TSF-B) \cite{bertasius2021space} is adopted as the visual encoder to extract visual representation, and a 12-layer Transformer ~\cite{vaswani2017attention} is used to obtain textual representation. The context length of the text is kept at most 77 tokens. Moreover, we also use a larger version of LAVILA-L (TSF-L as the visual encoder) for fair comparison with state-of-the-art (SOTA) methods. \par
\textbf{EgoTP$\spadesuit$}: For a $K$-category dataset, we expand semantics of each category $\textit{[C]}$ by command ChatGPT to generate diverse textual descriptions following four series of ego-oriented text prompts: \ding{172} Scene Description: ``{\color{gray!120}\textit{Describe the egocentric action of [C] in detail}}"; \ding{173} Question Answering: ``{{\color{gray!120}\textit{How can you identify the action of [C] in an egocentric video?}}"; \ding{174} Action Associating: ``{{{\color{gray!120}\textit{What does the egocentric action of [C] look like?}}"; \ding{175} Caption Generation: ``{{\color{gray!120}\textit{A caption of an egocentric action of [C]}:}}". These descriptions are then to output text representation after encoding by the textual encoder. The dimension of text representation is set to 768. For a $K$-category dataset, the output text descriptions are $\mathcal{T}_{des} \in \mathbb{R}^{K \times 768}$. \par
\textbf{EgoVP$\clubsuit$}: For a input video, we randomly sample $T=16$ frames to construct input video clip $\bm{X} \in \mathbb{R}^{16 \times 3 \times 224 \times 224}$. Moreover, the resolution of the frame is also adapted to $336 \times 336$ for fair comparison with SOTA methods. For each frame in input video clip $\bm{X}$, we utilize SAM~\cite{SAM} to segment $N$ pixel-level masks of refined visual concepts. The number of $N$ is different in different frames according to the degree of semanticity in each frame. We set the hyper-parameter $Q$ to make a quantitative selection to obtain the {best} performance. In Scene Parsing and Concept Augmentation, the number of $Q$ is set as 8 by investigating $Q \in \left\{ 0, 4, 8, 16\right\}$ in experiments. For an input video clip ${\bm{X}}$, the output video representation is $\mathcal{T}_{des} \in \mathbb{R}^{1 \times 768}$ after encoding by visual encoder.

\begin{table*}[!t]
    \caption{Comparison with state-of-the-art methods on three egocentric benchmarks, i.e., EK100, EGTEA, and CharadesEgo. Here, top-1/5 (\%), mAP (\%), mean acc. (\%) denote top-1/5 accuracy, mean average precision, and mean of the {accuracy} of each action category, respectively. $\bigtriangleup$$_{\rm{GPT4Ego^{B/L}-LAVILA^{B/L}}}$ denotes the performance gain from LAVILA-B/L to GPT4Ego-B/L.
}
\setlength\tabcolsep{5pt}
\centering
\scriptsize
\begin{tabular}{c|c|c|c|c|c|c|c|c|c}
    \hline
    \hline
    \multirow{2}*{  Method } & \multirow{2}*{  Venue } & \multirow{2}*{  Input } & \multirow{2}*{  Backbone } & \multirow{2}*{ Pre-training } & {CharadesEgo} & \multicolumn{2}{c}{EK100} & \multicolumn{2}{|c}{EGTEA} \\
    \cline{6-10}
    ~ & & & & & mAP (\%) & top-1 (\%) & top-5 (\%) & mean acc. (\%) & top-1 (\%) \\
    \hline 
    HierVL-Avg \cite{HierVL_2023_CVPR} & CVPR'23 & $16\times224^{2}$ & {\tiny{FrozenInTime}} & Ego4D & 25.2 & -  & - & - & -  \\
    HierVL-SA \cite{HierVL_2023_CVPR} & CVPR'23 & $16\times224^{2}$ & {\tiny{FrozenInTime}} & Ego4D & 26.0 & - & - & - & -  \\
    EgoVLP-Frozen \cite{egovlp_NeurIPS2022} & NeurIPS'22 & $16\times224^{2}$ &TSF-B & EgoCLIP & 23.6 & -  & - & - & -  \\
    EgoVLP-EgoNCE \cite{egovlp_NeurIPS2022} & NeurIPS'22 & $16\times224^{2}$ &TSF-B & EgoCLIP & 25.0 & -  & - & - & -  \\    
    EgoVLPv2 \cite{EgoVLPv2_ICCV2023} &ICCV"23 & $16\times224^{2}$ &TSF-B & EgoCLIP & 26.2 & -  & - & - & -  \\    
    LAVILA-B \cite{lavila_2023_CVPR} & CVPR'23 & $16\times224^{2}$ &TSF-B & WIT+Ego4D & 26.8 & 16.3 & 34.4 & 28.9 & 35.5  \\
    LAVILA-L \cite{lavila_2023_CVPR} & CVPR'23 & $16\times336^{2}$ &TSF-L & WIT+Ego4D & 28.9 & 23.8 & 48.2 & 34.1  & 41.9  \\
    \rowcolor{gray!20} \textbf{GPT4Ego-B} (ours) & - & $16\times224^{2}$ &TSF-B & WIT+Ego4D & 29.6 & 28.9 & 49.7 & 34.9 & 42.6  \\
\rowcolor{gray!20} \textbf{GPT4Ego-L} (ours) & - & $16\times336^{2}$ &TSF-L & WIT+Ego4D & 31.5 & 33.2 & 56.0 & 39.6 & 48.6   \\
    \hline {\color{blue}\small$\bigtriangleup$}$_{\color{blue}\rm{GPT4Ego^{B}-LAVILA^{B}}}$ & - & $16\times224^{2}$ & TSF-B & WIT+Ego4D & {\color{teal}+2.8} & {\color{teal}+12.6} & {\color{teal}+15.3} & {\color{teal}+6.0}  & {\color{teal}+7.1}  \\
    {\color{blue}\small$\bigtriangleup$}$_{\color{blue}\rm{GPT4Ego^{L}-LAVILA^{L}}}$ & - & $16\times336^{2}$ & TSF-L & WIT+Ego4D & {\color{teal}+2.6} & {\color{teal}+9.4} & {\color{teal}+7.8} & {\color{teal}+5.5}  & {\color{teal}+6.7}  \\
\hline
\hline
\end{tabular}
\label{sota_compare}
\end{table*}

\subsection{Comparison with State-of-the-arts}
\label{compare sota}

Table{\color{red}~\ref{sota_compare}} shows the performance comparison between the proposed GPT4Ego and the state-of-the-art (SOTA) methods on three egocentric benchmarks, i.e., EK100 \cite{damen2020rescaling}, EGTEA \cite{egtea}, and CharadesEgo \cite{Charades-ego}. Here, top-1/5 (\%), mAP (\%), mean acc. (\%) denote top-1/5 accuracy, mean average precision, and mean of the precision of each action category, respectively. Experimental results demonstrate that GPT4Ego achieves the best performance and outperforms other SOTA methods by a large margin on all egocentric benchmarks.  

Among these methods, EgoVLP \cite{egovlp_NeurIPS2022} and EgoVLP v2 \cite{EgoVLPv2_ICCV2023} focus on egocentric video-language pretraining. EgoVLP creates a new egocentric dataset named EgoClip well-chosen from Ego4D, a pretraining objective called EgoNCE for video-text contrastive learning, and a development benchmark called EgoMCQ for validating design decisions in EgoNCE and EgoClip. EgoVLP v2, the second generation of EgoVLP, highlights the cross-modal fusion and efficient pre-training. LAVILA \cite{lavila_2023_CVPR} proposes an approach to learning video-language representations by leveraging Large Language Models (GPT-2), which first densely narrate long videos by LLM, and then use those narrations to pre-train video models. HierVL introduces hierarchical video-language embedding to simultaneously learn both long-term and short-term associations.

\begin{table}[!t]
\caption{Performance of each GPT4Ego component on EK100, where $\bigtriangleup$(\%) shows gains over Base, with the best result highlighted.}

\centering
\begin{tabular}{l|c|c}
\hline
\hline
\text{Methods} & \text {Top-1 (\%)} & \text {$\bigtriangleup$ (\%)} \\
\hline \text {A0 (Base)} & {16.3} & {-} \\
\text  {A1 (Base + EgoTP$\spadesuit$)}  & {26.7} & {\color{teal}+10.4} \\
\text  {A2 (Base + EgoVP$\clubsuit$)} & {18.3} & {\color{teal}+2.0} \\
\text  {A3 (Base + EgoTP$\spadesuit$ + EgoVP$\clubsuit$)} & {\textbf{28.9}} & {\color{teal}+12.6} \\
\hline
\hline
\end{tabular}
\label{commpomt_effect}
\end{table}

\begin{table}[!t]
\caption{Effect of different LLMs in GPT4Ego on CharadesEgo, where $\bigtriangleup$(\%) shows gains over Base, with the best result highlighted.}

\centering
\begin{tabular}{l|c|>{\color{teal}}c}
\hline
\hline
\text{Versions} & \text {mAP (\%)} & \text {\color{black}$\bigtriangleup$ (\%)} \\
\hline \text {B0 (Base)} & {26.8} & {-} \\
\text {B1 (GPT-3)} & {27.1} & {+0.3} \\
\text {B2 (GPT-3.5)}  & {27.6} & {+0.8} \\
\text {B3 (ChatGPT)} & {\textbf{29.6}} & {+2.8} \\
\text {B4 (GPT-4)} & {28.1} & {+1.3} \\
\hline
\hline
\end{tabular}
\label{GPT_Versions}

\end{table}

Notably, compared with the LAVILA \cite{lavila_2023_CVPR} with the currently best performance, our GPT4Ego is still significantly ahead of it, when the input, backbone, and pre-training are fixed. Specifically, take LAVILA-B as an example, our GPT4Ego-B (TSF-B as the backbone) outperforms LAVILA: 1) EK100 by +12.6\% (top-1) and +15.3\% ( top-5); 2) EGTEA by +6.0\% (mean acc.) and +7.1\% (top-1); 3) CharadesEgo by +2.8\% (mAP). It is demonstrated that GPT4Ego can effectively improve the zero-shot performance.

\begin{figure*}[t!]
    \centering
    \includegraphics[width=\linewidth]{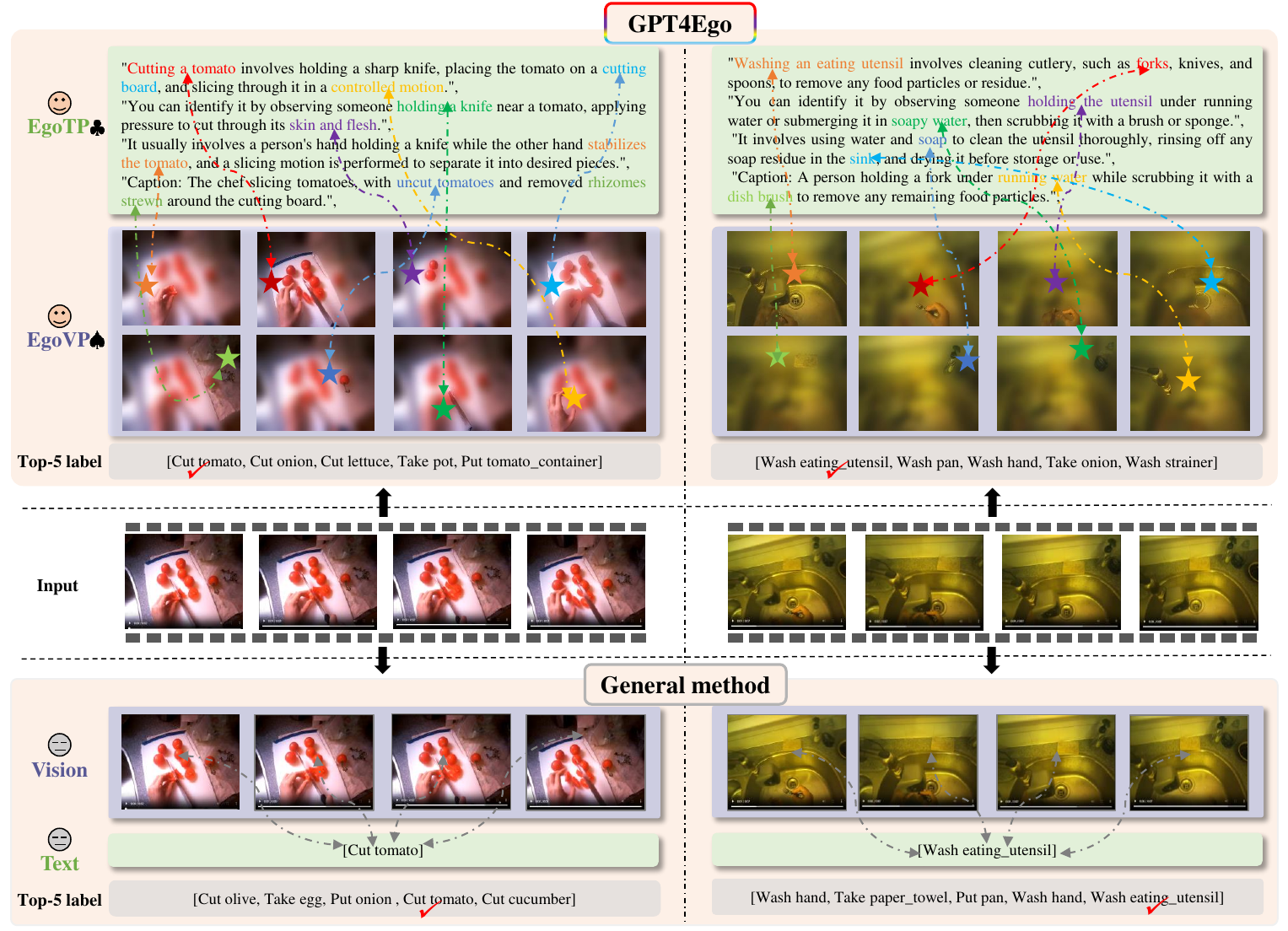}
    \caption{Visualization of textual descriptions and visual concepts generated by GPT4Ego. For a better view, the fine-grained description-concept alignment in vision and language is explicitly highlighted with multi-color lines. For comparison, we report the predictions of the top-5 class labels obtained by GPT4Ego and the general method.}
    \label{visualtion}
\end{figure*}

\subsection{Ablation Studies}
\label{Ablation Studies}
To illustrate the superior performance of GPT4Ego, we perform sets of ablation studies, including verifying the effect of each component, GPT versions, text prompt, and the number of visual concepts. The details are as follows.

\begin{itemize}
    \item \textbf{\textit{Effect of each component in GPT4Ego.}} We conduct an ablation study to validate the superiority of the main components in GPT4Ego on EK100. Table{\color{red}~\ref{commpomt_effect}} shows the performance of GPT4Ego with different components, i.e., EgoTP$\spadesuit$ and EgoVP$\clubsuit$. Here, A0 (Base) denotes that only uses the baseline (\textit{i.e.}, Lavila-B) model for zero-shot action recognition. Compared with A0, the performance of A1 (Base + EgoTP$\spadesuit$) is significantly improved, by improving +10.4 (\%) in top-1 accuracy. It proves that EgoTP$\spadesuit$ can effectively enhance the performance of action recognition by prompting rich textual descriptions.

\begin{table}[!t]

\caption{Effect of each text prompt in EgoTP$\spadesuit$ on CharadesEgo, where $\bigtriangleup$(\%) shows gains over Base, with the best result highlighted.}

\centering
\begin{tabular}{l|c|c}
\hline
\hline
\text{Prompts} & \text { mAP (\%) } & \text {$\bigtriangleup$ (\%)} \\
\hline \text {C0 (Base)} & {26.8} & {-} \\
\text {C1 (Base + Prompt\ding{172})} & {28.2} & {\color{teal}+1.4} \\
\text {C2 (Base + Prompt\ding{173})} & {27.6} & {\color{teal}+0.8} \\
\text {C3 (Base + Prompt\ding{174})} & {28.0} & {\color{teal}+1.2} \\
\text {C4 (Base + Prompt\ding{175})} & {27.7} & {\color{teal}+0.9} \\
\text {C5 (\textbf{GPT4Ego)}} & {\textbf{29.6}} & {\color{teal}+2.8} \\
\hline
\hline
\end{tabular}
\label{each_text_prompt}

\end{table}

\begin{table}[!t]

\caption{Impact of visual concept quantities in EgoVP$\clubsuit$ on EK100.}

\centering
\begin{tabular}{l|c}
\hline
\hline
\text{Concepts} & \text {Top-1 (\%)} \\
\hline D0 (Base) & {16.3} \\
\text D1 ($Q$=2) & {26.8} \\
\text D2 ($Q$=4) & {27.2} \\
\text D3 ($Q$=8) & {28.9} \\
\text D4 ($Q$=16) & {29.0} \\
\hline
\hline
\end{tabular}
\label{cocepts_table}

\end{table}

\begin{figure*}[t!]
    \centering
    \includegraphics[width=\linewidth]{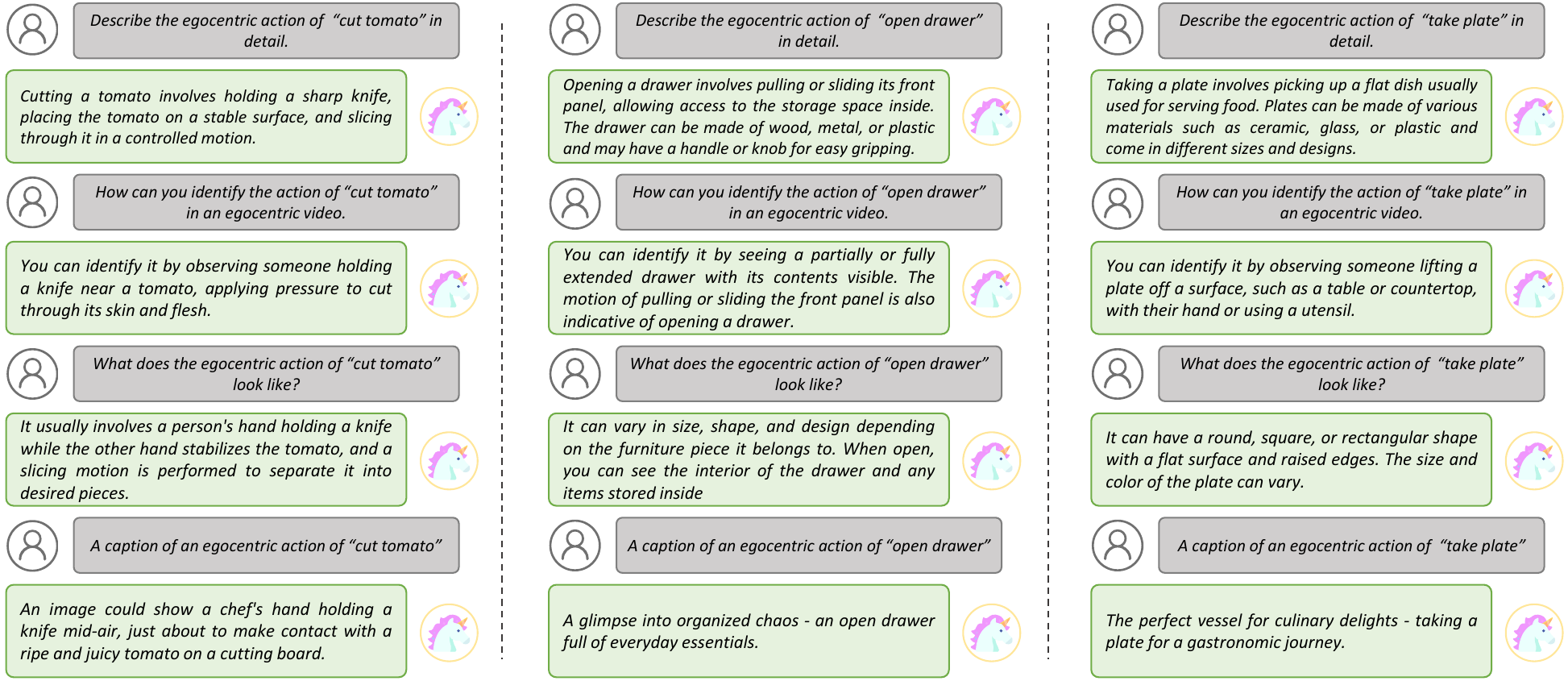}
    \caption{Visualization of the qualitative results of interacting with ChatGPT by following the chain-of-thought design prompts in GPT4Ego. Regarding each example, for a better view, the outcomes of the textual prompts and ChatGPT responses are denoted by \logouser \ and\ \logorespomse, respectively.}
    \label{chat_log}
\end{figure*}

\begin{figure*}[t!]
    \centering
    \includegraphics[width=\linewidth]{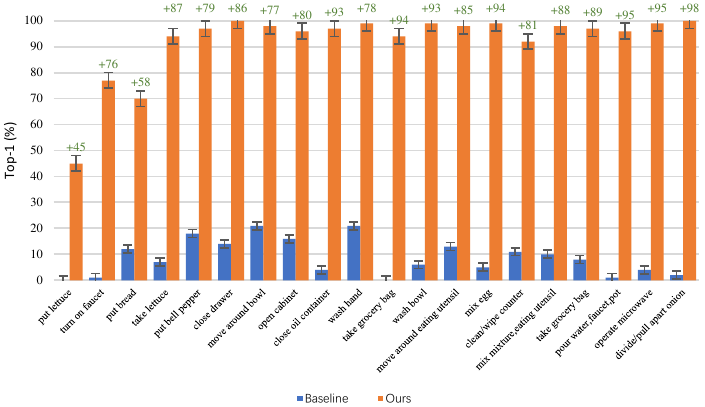}
    \caption{Visualization of improvements in the top-1 accuracy of some categories of our method compared to the baseline method. For the sake of view, the magnitude of the performance gains with the same category is explicitly highlighted in {\color{teal} \textbf{green}}.}
    \label{categray_top}
\end{figure*}

Compared with {A0}, the performance gain of A2 is continued when equipped with EgoVP$\clubsuit$, which validates the effectiveness of refined vision-contextual semantics. Moreover, the above experimental results also demonstrate that either EgoTP$\spadesuit$ or EgoVP$\clubsuit$ can also create a great accuracy gain for ZS-EAR. Thus, it motivates us to validate the effect of using EgoTP$\spadesuit$ and EgoVP$\clubsuit$ together. From the results in A3, when we combine EgoTP$\spadesuit$ and EgoVP$\clubsuit$ 
with Base, it further improves the recognition accuracy on a large margin, by +12.6 (\%) in top-1 accuracy. To sum up, all components in GPT4Ego can jointly improve the performance. \par

    \item \textbf{\textit{Effect of GPT versions in GPT4Ego.}} We conduct an ablation study to validate the effect of using different GPT versions in GPT4Ego on CharadesEgo. In this work, we validate a total of four GPT versions following the order of their release by Openen AI, and the performance results are shown in Table{\color{red}~\ref{GPT_Versions}}. The performance of ZS-EAR is gradually increasing with the updating of the GPT version in addition to ChatGPT. In Table~\ref{GPT_Versions}, ChatGPT achieves the best performance. This may be because that ChatGPT is more specialized in command dialogue and context understanding than the other GPT versions. Therefore, in GPT4Ego, we finally chose ChatGPT to generate diverse textual descriptions. \par
    
    \item \textbf{\textit{Effect of each text prompt (EgoTP$\spadesuit$) in GPT4Ego.}} We conduct an ablation study to validate the effect of each prompt in GPT4Ego on CharadesEgo, and the results are shown in Table{\color{red}~\ref{each_text_prompt}}. Here, C0 (Base) denotes the baseline model, and Prompt\ding{172}-\ding{175} denote only use no.1-no.4 textual prompts, denoted by B1-B4, respectively. Compared with the base model, the performances of B1-B4 are all improved and the first textual prompt (Prompt\#1) improved the most, by +1.4\% in top-1 accuracy. It demonstrates that Prompt\ding{172}, i.e. Scene Description: ``\textit{\color{gray!120}{{Describe the egocentric action of [C] in detail}}}", can provide more {text-contextual information} than other textual prompts, effectively aiding fine-grained alignment with visual features. Finally, the integration of all textual prompts in B5 (GPT4Ego) yields the highest accuracy at 29.6\%. \par
    
    \item \textbf{\textit{Effect of numbers of visual concepts (EgoVP$\clubsuit$) in GPT4Ego.}} We conduct the ablation study to investigate the number $Q$ of refined visual concepts in GPT4Ego on EK100. Table{\color{red}~\ref{cocepts_table}} shows the performance of GPT4Ego with different values of $Q$. Here, D0 (Base) denotes the baseline model, which only uses the original frame for zero-shot action recognition. We set $Q\in\{2,4,8,16\}$ in D1-D4 for investigation, and select $Q=8$ when the best performance is achieved. It can be found that when $Q$ is set to smaller than 8 or larger than 8, the accuracy of zero-shot recognition decreases slightly. Nevertheless, a larger or smaller number of visual-contextual concepts in a moderate range is acceptable since refined concepts are generated by SAM in an automated way.   
\end{itemize} 

{With the above-mentioned ablation experiments, we demonstrate the effectiveness of each component of the proposed method. In addition, it also proves that the proposed method is available and credible by further investigating the detailed parameters in each component.}

\subsection{Qualitative Analysis}
\label{Qualitative Analysis}
{In this section, we perform sets of visualization analyses to illustrate the qualitative performance of GPT4Ego as follows.} 
\begin{itemize}
    \item \textbf{\textit{Visualization of fine-grained semantics matching.}} To validate the effect of GPT4Ego for effectively matching fine-grained semantics in vision and language, we report the visualization comparison between fine-grained concept-description alignment in our GPT4Ego and the coarse-grained video-text alignment in the general VLM-based method as shown in Figure{\color{red}~\ref{visualtion}}. Moreover, we also report top-5 predictions for comparison. Here, the lines with multiple colors denote the explicit co-existing semantics in vision and language, and the corrected predictions of {logits} are marked with a checkmark. Take the action ``{\texttt{Cut tomato}}'' as an example, for the same input video clip, the contextual semantics explicitly captured by GPT4Ego are more than those by the general method, leading to more accurate class prediction. This proves the effect of fine-grained {concept-description} alignment, improving the performance of ZS-EAR. \par
    \item \textbf{\textit{Visualization of dialog response of designed prompts.}} In order to illustrate the effect of the designed prompts in GPT4Ego, a imitate the idea of chain-of-thought, we visualize the response of these prompts after interacting with ChatGPT step-by-step, as shown in Figure{\color{red}~\ref{chat_log}}. Take the action ``open drawer" as an example, GPT4Ego imitates human thinking to first gain a holistic understanding of the action, then summarize the steps of how to perform it, then make associations to enrich the understanding, and finally provide a descriptive summary of the action. Through this process, GPT4Ego effectively migrates the ``open drawer" action understanding from the GPT step by step using stimulus thinking and thus obtains a richer and more comprehensive semantic understanding. \par
    \item \textbf{\textit{Visualization of improvements for hard categories.}} To better view the details of the proposed GPT4Ego for performance improvements, we report top-1 accuracy on EGTEA for ranking the top 20 hard categories, where the baseline model recognizes them have poor performance. Figure~{\color{red}\ref{categray_top}} showcases the comparison results, our GPT4Ego significantly outperforms the baseline method in these categories. Our conjecture is that naive category names may convey more global concepts, whereas the more comprehensive details in the GPT4Ego by generating action-related text-contextual semantics and vision-contextual semantics may be closer to the video content, thus enhancing inter-class variability.
\end{itemize}

\section{Conclusion and Discussion}
\label{Conclusion}
\textbf{Conclusion.} We introduce GPT4Ego, a simple yet effective VLM framework for improving Zero-Shot Egocentric Action Recognition (ZS-EAR), which pursues fine-grained semantic alignment between vision and language by prompting more visual concepts and textual descriptions as the contextual semantics. GPT4Ego has two insightful modules, i.e., Ego-oriented Text Prompting (EgoTP$\spadesuit$), and Ego-oriented Visual Parsing (EgoVP$\clubsuit$). For improving action-related text-contextual semantics, EgoTP$\spadesuit$ learns diverse text representation by evolving word-level class names to sentence-level contextual descriptions by ChatGPT with well-designed textual prompts. For improving action-related vision-contextual semantics, EgoVP$\clubsuit$ learns rich visual representations by refining global-level images to part-level contextual concepts with the help of SAM. To the best of our knowledge, GPT4Ego is the first work that integrates SAM and GPT into VLMs for promoting the semantic alignment between vision and language for ZS-EAR without any tuning. Extensive experiments on three egocentric video benchmarks demonstrate that GPT4Ego significantly outperforms the state-of-the-arts. 

\textbf{Discussion.} For the proposed GPT4Ego framework, this is a new paradigm that focus on the fine-grained semantic alignment between vision and language in VLMs besides just rough alignment. Both EgoTP$\spadesuit$ and EgoVP$\clubsuit$ are plug-and-play for enforcing textual and visual contextual semantics. This well matches our intention, contributing feasible modules like EgoTP$\spadesuit$ and EgoVP$\clubsuit$ in the video understanding community. One more thing, our framework is very simple though we can still add a lot of technological components for further performance improvement. Nevertheless, like CLIP, what we pursue is ``{{Less}} is {{More}}".

\bibliographystyle{IEEEtran}
\bibliography{reference}

\begin{thebibliography}{100}

\bibitem{Alayrac2022Flamingo}
Jean-Baptiste Alayrac, Jeff Donahue, Pauline Luc, Antoine Miech, Iain Barr, Yana Hasson, Karel Lenc, Arthur Mensch, Katherine Millican, Malcolm Reynolds, and et~al.
\newblock Flamingo: a visual language model for few-shot learning.
\newblock In {\em Advances in Neural Information Processing Systems (NeurIPS)}, pages 23716--23736, 2022.

\bibitem{GPT-2}
Radford Alec, Wu~Jeffrey, Child Rewon, Luan David, Amodei Dario, and et~al.
\newblock Language models are unsupervised multitask learners.
\newblock In {\em OpenAI blog}, page~9, 2019.

\bibitem{arnab2021vivit}
Anurag Arnab, Mostafa Dehghani, Georg Heigold, Chen Sun, Mario Lu{\v{c}}i{\'c}, and Cordelia Schmid.
\newblock Vivit: A video vision transformer.
\newblock In {\em Proceedings of the IEEE/CVF International Conference on Computer Vision (ICCV)}, pages 6836--6846, 2021.

\bibitem{HierVL_2023_CVPR}
Kumar Ashutosh, Rohit Girdhar, Lorenzo Torresani, and Kristen Grauman.
\newblock Hiervl: Learning hierarchical video-language embeddings.
\newblock In {\em Proceedings of the IEEE/CVF Conference on Computer Vision and Pattern Recognition (CVPR)}, pages 23066--23078, 2023.

\bibitem{bertasius2021space}
Gedas Bertasius, Heng Wang, and Lorenzo Torresani.
\newblock Is space-time attention all you need for video understanding?
\newblock In {\em Proceedings of the International Conference on Machine Learning (ICML)}, pages 813--824, 2021.

\bibitem{GPT-3}
Tom~B. Brown, Benjamin Mann, Nick Ryder, Melanie Subbiah, Jared Kaplan, and et~al.
\newblock Language models are few-shot learners.
\newblock In {\em Advances in Neural Information Processing Systems (NeurIPS)}, pages 1877--1901, 2020.

\bibitem{bulat2021space}
Adrian Bulat, Juan~Manuel Perez~Rua, Swathikiran Sudhakaran, Brais Martinez, and Georgios Tzimiropoulos.
\newblock Space-time mixing attention for video transformer.
\newblock In {\em Advances in Neural Information Processing Systems (NeurIPS)}, pages 19594--19607, 2021.

\bibitem{calway2015discovering}
A~Calway, W~Mayol-Cuevas, D~Damen, O~Haines, and T~Leelasawassuk.
\newblock Discovering task relevant objects and their modes of interaction from multi-user egocentric video.
\newblock In {\em Proceedings of the British Machine Vision Conference (BMVC)}, pages 1--13, 2015.

\bibitem{carreira2017quo}
Joao Carreira and Andrew Zisserman.
\newblock Quo vadis, action recognition? a new model and the kinetics dataset.
\newblock In {\em Proceedings of the IEEE/CVF Conference on Computer Vision and Pattern Recognition (CVPR)}, pages 6299--6308, 2017.

\bibitem{chen2020simple}
Ting Chen, Simon Kornblith, Mohammad Norouzi, and Geoffrey Hinton.
\newblock A simple framework for contrastive learning of visual representations.
\newblock In {\em Proceedings of the International Conference on Machine Learning (ICML)}, pages 1597--1607, 2020.

\bibitem{PALM}
Aakanksha Chowdhery, Sharan Narang, Jacob Devlin, Maarten Bosma, Gaurav Mishra, and et~al.
\newblock Palm: Scaling language modeling with pathways.
\newblock {\em arXiv preprint arXiv:2204.02311}, 2022.

\bibitem{damen2018scaling}
Dima Damen, Hazel Doughty, Giovanni~Maria Farinella, Sanja Fidler, Antonino Furnari, Evangelos Kazakos, Davide Moltisanti, Jonathan Munro, Toby Perrett, Will Price, et~al.
\newblock Scaling egocentric vision: The epic-kitchens dataset.
\newblock In {\em Proceedings of the European Conference on Computer Vision (ECCV)}, pages 720--736, 2018.

\bibitem{Damen2020EPIC-KITCHENS-55}
Dima Damen, Hazel Doughty, Giovanni~Maria Farinella, Sanja Fidler, Antonino Furnari, Evangelos Kazakos, Davide Moltisanti, Jonathan Munro, Toby Perrett, Will Price, and Michael Wray.
\newblock The epic-kitchens dataset: Collection, challenges and baselines.
\newblock {\em IEEE Transactions on Pattern Analysis and Machine Intelligence}, 43(11):4125--4141, 2021.

\bibitem{damen2020rescaling}
Dima Damen, Hazel Doughty, Giovanni~Maria Farinella, Antonino Furnari, Evangelos Kazakos, Jian Ma, Davide Moltisanti, Jonathan Munro, Toby Perrett, Will Price, et~al.
\newblock Rescaling egocentric vision.
\newblock {\em arXiv preprint arXiv:2006.13256}, 2020.

\bibitem{damen2014you}
Dima Damen, Teesid Leelasawassuk, Osian Haines, Andrew Calway, and Walterio~W Mayol-Cuevas.
\newblock You-do, i-learn: Discovering task relevant objects and their modes of interaction from multi-user egocentric video.
\newblock In {\em Proceedings of the British Machine Vision Conference (BMVC)}, pages 1--13, 2014.

\bibitem{del2016summarization}
Ana~Garcia Del~Molino, Cheston Tan, Joo-Hwee Lim, and Ah-Hwee Tan.
\newblock Summarization of egocentric videos: A comprehensive survey.
\newblock {\em IEEE Transactions on Human-Machine Systems}, 47(1):65--76, 2016.

\bibitem{vln_img_seg}
Jian Ding, Nan Xue, Gui-Song Xia, Bernt Schiele, and Dengxin Dai.
\newblock Hgformer: Hierarchical grouping transformer for domain generalized semantic segmentation.
\newblock In {\em Proceedings of the IEEE/CVF Conference on Computer Vision and Pattern Recognition (CVPR)}, pages 15413--15423, 2023.

\bibitem{GPT4Image}
Ning Ding, Yehui Tang, Zhongqian Fu, Chao Xu, Kai Han, and Yunhe Wang.
\newblock Gpt4image: Can large pre-trained models help vision models on perception tasks?
\newblock {\em arXiv preprint arXiv:2306.00693}, 2023.

\bibitem{dong2022cswin}
Xiaoyi Dong, Jianmin Bao, Dongdong Chen, Weiming Zhang, Nenghai Yu, Lu~Yuan, Dong Chen, and Baining Guo.
\newblock Cswin transformer: A general vision transformer backbone with cross-shaped windows.
\newblock In {\em Proceedings of the IEEE/CVF Conference on Computer Vision and Pattern Recognition (CVPR)}, pages 12124--12134, 2022.

\bibitem{dosovitskiy2020image}
Alexey Dosovitskiy, Lucas Beyer, Alexander Kolesnikov, Dirk Weissenborn, Xiaohua Zhai, Thomas Unterthiner, Mostafa Dehghani, Matthias Minderer, Georg Heigold, Sylvain Gelly, et~al.
\newblock An image is worth 16x16 words: Transformers for image recognition at scale.
\newblock {\em arXiv preprint arXiv:2010.11929}, 2020.

\bibitem{escorcia2022sos}
Victor Escorcia, Ricardo Guerrero, Xiatian Zhu, and Brais Martinez.
\newblock Sos! self-supervised learning over sets of handled objects in egocentric action recognition.
\newblock In {\em Proceedings of the European Conference on Computer Vision (ECCV)}, pages 604--620, 2022.

\bibitem{fan2021multiscale}
Haoqi Fan, Bo~Xiong, Karttikeya Mangalam, Yanghao Li, Zhicheng Yan, Jitendra Malik, and Christoph Feichtenhofer.
\newblock Multiscale vision transformers.
\newblock In {\em Proceedings of the IEEE/CVF International Conference on Computer Vision (ICCV)}, pages 6824--6835, 2021.

\bibitem{fang2021you}
Yuxin Fang, Bencheng Liao, Xinggang Wang, Jiemin Fang, Jiyang Qi, Rui Wu, Jianwei Niu, and Wenyu Liu.
\newblock You only look at one sequence: Rethinking transformer in vision through object detection.
\newblock In {\em Advances in Neural Information Processing Systems (NeurIPS)}, pages 26183--26197, 2021.

\bibitem{fathi2012learning}
Alireza Fathi, Yin Li, James~M Rehg, et~al.
\newblock Learning to recognize daily actions using gaze.
\newblock In {\em Proceedings of the European Conference on Computer Vision (ECCV)}, pages 314--327, 2012.

\bibitem{feichtenhofer2019slowfast}
Christoph Feichtenhofer, Haoqi Fan, Jitendra Malik, and Kaiming He.
\newblock Slowfast networks for video recognition.
\newblock In {\em Proceedings of the IEEE/CVF International Conference on Computer Vision (ICCV)}, pages 6202--6211, 2019.

\bibitem{tc4}
Jie Fu, Junyu Gao, and Changsheng Xu.
\newblock Learning semantic-aware spatial-temporal attention for interpretable action recognition.
\newblock {\em IEEE Transactions on Circuits and Systems for Video Technology}, 32(8):5213--5224, 2022.

\bibitem{tc5}
Jie Fu, Junyu Gao, and Changsheng Xu.
\newblock Learning semantic-aware spatial-temporal attention for interpretable action recognition.
\newblock {\em IEEE Transactions on Circuits and Systems for Video Technology}, 32(8):5213--5224, 2022.

\bibitem{furnari2018leveraging}
Antonino Furnari, Sebastiano Battiato, and Giovanni Maria~Farinella.
\newblock Leveraging uncertainty to rethink loss functions and evaluation measures for egocentric action anticipation.
\newblock In {\em Proceedings of the European Conference on Computer Vision Workshops (ECCVW)}, 2018.

\bibitem{furnari2020rolling}
Antonino Furnari and Giovanni~Maria Farinella.
\newblock Rolling-unrolling lstms for action anticipation from first-person video.
\newblock {\em IEEE Transactions on Pattern Analysis and Machine Intelligence}, 43(11):4021--4036, 2020.

\bibitem{girdhar2022omnivore}
Rohit Girdhar, Mannat Singh, Nikhila Ravi, Laurens van~der Maaten, Armand Joulin, and Ishan Misra.
\newblock Omnivore: A single model for many visual modalities.
\newblock In {\em Proceedings of the IEEE/CVF Conference on Computer Vision and Pattern Recognition (CVPR)}, pages 16102--16112, 2022.

\bibitem{ego4d_cvpr2022}
Kristen Grauman, Andrew Westbury, Eugene Byrne, Zachary Chavis, Antonino Furnari, and et~al.
\newblock Ego4d: Around the world in 3,000 hours of egocentric video.
\newblock In {\em Proceedings of the IEEE/CVF Conference on Computer Vision and Pattern Recognition (CVPR)}, pages 18995--19012, 2022.

\bibitem{grauman2022ego4d}
Kristen Grauman, Andrew Westbury, Eugene Byrne, Zachary Chavis, Antonino Furnari, Rohit Girdhar, Jackson Hamburger, Hao Jiang, Miao Liu, Xingyu Liu, et~al.
\newblock Ego4d: Around the world in 3,000 hours of egocentric video.
\newblock In {\em Proceedings of the IEEE/CVF Conference on Computer Vision and Pattern Recognition (CVPR)}, pages 18995--19012, 2022.

\bibitem{Ego-Exo4Dgrauman2023ego}
Kristen Grauman, Andrew Westbury, Lorenzo Torresani, Kris Kitani, Jitendra Malik, Triantafyllos Afouras, Kumar Ashutosh, Vijay Baiyya, Siddhant Bansal, Bikram Boote, et~al.
\newblock Ego-exo4d: Understanding skilled human activity from first-and third-person perspectives.
\newblock {\em arXiv preprint arXiv:2311.18259}, 2023.

\bibitem{egotv}
Rishi Hazra, Brian Chen, Akshara Rai, Nitin Kamra, and Ruta Desai.
\newblock Egotv: Egocentric task verification from natural language task descriptions.
\newblock In {\em Proceedings of the IEEE/CVF International Conference on Computer Vision (ICCV)}, pages 15417--15429, 2023.

\bibitem{he2022masked}
Kaiming He, Xinlei Chen, Saining Xie, Yanghao Li, Piotr Doll{\'a}r, and Ross Girshick.
\newblock Masked autoencoders are scalable vision learners.
\newblock In {\em Proceedings of the IEEE/CVF Conference on Computer Vision and Pattern Recognition (CVPR)}, pages 16000--16009, 2022.

\bibitem{herzig2022object}
Roei Herzig, Elad Ben-Avraham, Karttikeya Mangalam, Amir Bar, Gal Chechik, Anna Rohrbach, Trevor Darrell, and Amir Globerson.
\newblock Object-region video transformers.
\newblock In {\em Proceedings of the IEEE/CVF Conference on Computer Vision and Pattern Recognition (CVPR)}, pages 3148--3159, 2022.

\bibitem{tmm1}
Yi~Huang, Xiaoshan Yang, Junyun Gao, and Changsheng Xu.
\newblock Holographic feature learning of egocentric-exocentric videos for multi-domain action recognition.
\newblock {\em IEEE Transactions on Multimedia}, 24:2273--2286, 2022.

\bibitem{mm22ear}
Yi~Huang, Xiaoshan Yang, and Changsheng Xu.
\newblock Multimodal global relation knowledge distillation for egocentric action anticipation.
\newblock In {\em Proceedings of the ACM International Conference on Multimedia (ACM MM)}, pages 245--254, 2021.

\bibitem{huang2020mutual}
Yifei Huang, Minjie Cai, Zhenqiang Li, Feng Lu, and Yoichi Sato.
\newblock Mutual context network for jointly estimating egocentric gaze and action.
\newblock {\em IEEE Transactions on Image Processing}, 29:7795--7806, 2020.

\bibitem{ALIGN}
Chao Jia, Yinfei Yang, Ye~Xia, Yi{-}Ting Chen, Zarana Parekh, Hieu Pham, Quoc~V. Le, Yun{-}Hsuan Sung, Zhen Li, and Tom Duerig.
\newblock Scaling up visual and vision-language representation learning with noisy text supervision.
\newblock In {\em Proceedings of the 38th International Conference on Machine Learning, (ICML)}, pages 4904--4916, 2021.

\bibitem{Kapidis_2019_ICCV}
Georgios Kapidis, Ronald Poppe, Elsbeth van Dam, Lucas Noldus, and Remco Veltkamp.
\newblock Multitask learning to improve egocentric action recognition.
\newblock In {\em Proceedings of the IEEE/CVF International Conference on Computer Vision Workshops (ICCVW)}, 2019.

\bibitem{kay2017kinetics}
Will Kay, Joao Carreira, Karen Simonyan, Brian Zhang, Chloe Hillier, Sudheendra Vijayanarasimhan, Fabio Viola, Tim Green, Trevor Back, Paul Natsev, et~al.
\newblock The kinetics human action video dataset.
\newblock {\em arXiv preprint arXiv:1705.06950}, 2017.

\bibitem{kazakos2021MTCN}
Evangelos Kazakos, Jaesung Huh, Arsha Nagrani, Andrew Zisserman, and Dima Damen.
\newblock With a little help from my temporal context: Multimodal egocentric action recognition.
\newblock In {\em Proceedings of the British Machine Vision Conference (BMVC)}, pages 1--16, 2021.

\bibitem{kazakos2019epic}
Evangelos Kazakos, Arsha Nagrani, Andrew Zisserman, and Dima Damen.
\newblock Epic-fusion: Audio-visual temporal binding for egocentric action recognition.
\newblock In {\em Proceedings of the IEEE/CVF International Conference on Computer Vision (ICCV)}, pages 5492--5501, 2019.

\bibitem{egohumans2023khirodkar}
Rawal Khirodkar, Aayush Bansal, Lingni Ma, Richard Newcombe, Minh Vo, and Kris Kitani.
\newblock Egohumans: An egocentric 3d multi-human benchmark.
\newblock {\em arXiv preprint arXiv:2305.16487}, 2023.

\bibitem{kim2022detector}
Dongkeun Kim, Jinsung Lee, Minsu Cho, and Suha Kwak.
\newblock Detector-free weakly supervised group activity recognition.
\newblock In {\em Proceedings of the IEEE/CVF Conference on Computer Vision and Pattern Recognition (CVPR)}, pages 20083--20093, 2022.

\bibitem{SAM}
Alexander Kirillov, Eric Mintun, Nikhila Ravi, Hanzi Mao, Chloe Rolland, Laura Gustafson, Tete Xiao, Spencer Whitehead, Alexander~C Berg, Wan-Yen Lo, et~al.
\newblock Segment anything.
\newblock {\em arXiv preprint arXiv:2304.02643}, 2023.

\bibitem{kuehne2011hmdb}
Hildegard Kuehne, Hueihan Jhuang, Est{\'\i}baliz Garrote, Tomaso Poggio, and Thomas Serre.
\newblock Hmdb: a large video database for human motion recognition.
\newblock In {\em Proceedings of the IEEE/CVF International Conference on Computer Vision (ICCV)}, pages 2556--2563, 2011.

\bibitem{lee2015predicting}
Yong~Jae Lee and Kristen Grauman.
\newblock Predicting important objects for egocentric video summarization.
\newblock {\em International Journal of Computer Vision}, 114:38--55, 2015.

\bibitem{li2022egocentric}
Haoxin Li, Wei-Shi Zheng, Jianguo Zhang, Haifeng Hu, Jiwen Lu, and Jian-Huang Lai.
\newblock Egocentric action recognition by automatic relation modeling.
\newblock {\em IEEE Transactions on Pattern Analysis and Machine Intelligence}, 45(1):489--507, 2022.

\bibitem{tc1}
Haoxin Li, Wei-Shi Zheng, Jianguo Zhang, Haifeng Hu, Jiwen Lu, and Jian-Huang Lai.
\newblock Motion stimulation for compositional action recognition.
\newblock {\em IEEE Transactions on Circuits and Systems for Video Technology}, 33(5):2061--2074, 2022.

\bibitem{blip-2}
Junnan Li, Dongxu Li, Silvio Savarese, and Steven Hoi.
\newblock Blip-2: Bootstrapping language-image pre-training with frozen image encoders and large language models.
\newblock {\em arXiv preprint arXiv:2301.12597}, 2023.

\bibitem{videochat}
KunChang Li, Yinan He, Yi~Wang, Yizhuo Li, Wenhai Wang, Ping Luo, Yali Wang, Limin Wang, and Yu~Qiao.
\newblock Videochat: Chat-centric video understanding.
\newblock {\em arXiv preprint arXiv:2305.06355}, 2023.

\bibitem{probatt2021}
Yin Li, Miao Liu, and Jame Rehg.
\newblock In the eye of the beholder: Gaze and actions in first person video.
\newblock {\em IEEE Transactions on Pattern Analysis and Machine Intelligence}, 2021.

\bibitem{egtea}
Yin Li, Miao Liu, and James~M Rehg.
\newblock In the eye of beholder: Joint learning of gaze and actions in first person video.
\newblock In {\em Proceedings of the European conference on computer vision (ECCV)}, pages 619--635, 2018.

\bibitem{li2023comprehensive}
Yingshu Li, Yunyi Liu, Zhanyu Wang, Xinyu Liang, Lingqiao Liu, Lei Wang, Leyang Cui, Zhaopeng Tu, Longyue Wang, and Luping Zhou.
\newblock A comprehensive study of gpt-4v's multimodal capabilities in medical imaging.
\newblock {\em medRxiv}, pages 2023--11, 2023.

\bibitem{lin2019tsm}
Ji~Lin, Chuang Gan, and Song Han.
\newblock Tsm: Temporal shift module for efficient video understanding.
\newblock In {\em Proceedings of the IEEE/CVF International Conference on Computer Vision (ICCV)}, pages 7083--7093, 2019.

\bibitem{lin2023mm}
Kevin Lin, Faisal Ahmed, Linjie Li, Chung-Ching Lin, Ehsan Azarnasab, Zhengyuan Yang, Jianfeng Wang, Lin Liang, Zicheng Liu, Yumao Lu, et~al.
\newblock Mm-vid: Advancing video understanding with gpt-4v (ision).
\newblock {\em arXiv preprint arXiv:2310.19773}, 2023.

\bibitem{egovlp_NeurIPS2022}
Kevin~Qinghong Lin, Alex~Jinpeng Wang, Mattia Soldan, Michael Wray, Rui Yan, Eric~Zhongcong Xu, Difei Gao, Rongcheng Tu, Wenzhe Zhao, Weijie Kong, et~al.
\newblock Egocentric video-language pretraining.
\newblock In {\em Advances in Neural Information Processing Systems (NeurIPS)}, pages 7575--7586, 2022.

\bibitem{liu2021wb}
Fanfan Liu, Haoran Wei, Wenzhe Zhao, Guozhen Li, Jingquan Peng, and Zihao Li.
\newblock Wb-detr: transformer-based detector without backbone.
\newblock In {\em Proceedings of the IEEE/CVF International Conference on Computer Vision (ICCV)}, pages 2979--2987, 2021.

\bibitem{liu2023hallusionbench}
Fuxiao Liu, Tianrui Guan, Zongxia Li, Lichang Chen, Yaser Yacoob, Dinesh Manocha, and Tianyi Zhou.
\newblock Hallusionbench: You see what you think? or you think what you see? an image-context reasoning benchmark challenging for gpt-4v (ision), llava-1.5, and other multi-modality models.
\newblock {\em arXiv preprint arXiv:2310.14566}, 2023.

\bibitem{liu2023llava}
Haotian Liu, Chunyuan Li, Qingyang Wu, and Yong~Jae Lee.
\newblock Visual instruction tuning, 2023.

\bibitem{tmm3}
Shaocan Liu and Xin Ma.
\newblock Attention-driven appearance-motion fusion network for action recognition.
\newblock {\em IEEE Transactions on Multimedia}, 25:2573--2584, 2023.

\bibitem{tc2}
Shuwen Liu, Min Jiang, and Jun Kong.
\newblock Multidimensional prototype refactor enhanced network for few-shot action recognition.
\newblock {\em IEEE Transactions on Circuits and Systems for Video Technology}, 32(10):6955--6966, 2022.

\bibitem{liu2021swin}
Ze~Liu, Yutong Lin, Yue Cao, Han Hu, Yixuan Wei, Zheng Zhang, Stephen Lin, and Baining Guo.
\newblock Swin transformer: Hierarchical vision transformer using shifted windows.
\newblock In {\em Proceedings of the IEEE/CVF International Conference on Computer Vision (ICCV)}, pages 10012--10022, 2021.

\bibitem{liu2022video}
Ze~Liu, Jia Ning, Yue Cao, Yixuan Wei, Zheng Zhang, Stephen Lin, and Han Hu.
\newblock Video swin transformer.
\newblock In {\em Proceedings of the IEEE/CVF Conference on Computer Vision and Pattern Recognition (CVPR)}, pages 3202--3211, 2022.

\bibitem{lu_lsae}
Minlong Lu, Danping Liao, and Ze-Nian Li.
\newblock Learning spatiotemporal attention for egocentric action recognition.
\newblock In {\em Proceedings of the IEEE/CVF International Conference on Computer Vision Workshops (ICCVW)}, pages 4425--4434, 2019.

\bibitem{lu2023mathvista}
Pan Lu, Hritik Bansal, Tony Xia, Jiacheng Liu, Chunyuan Li, Hannaneh Hajishirzi, Hao Cheng, Kai-Wei Chang, Michel Galley, and Jianfeng Gao.
\newblock Mathvista: Evaluating mathematical reasoning of foundation models in visual contexts.
\newblock {\em arXiv preprint arXiv:2310.02255}, 2023.

\bibitem{lyu2023gpt}
Hanjia Lyu, Jinfa Huang, Daoan Zhang, Yongsheng Yu, Xinyi Mou, Jinsheng Pan, Zhengyuan Yang, Zhongyu Wei, and Jiebo Luo.
\newblock Gpt-4v (ision) as a social media analysis engine.
\newblock {\em arXiv preprint arXiv:2311.07547}, 2023.

\bibitem{ma2022hand}
Jian Ma and Dima Damen.
\newblock Hand-object interaction reasoning.
\newblock In {\em Proceedings of the International Conference on Advanced Video and Signal Based Surveillance (AVSS)}, pages 1--8, 2022.

\bibitem{min2021integrating}
Kyle Min and Jason~J Corso.
\newblock Integrating human gaze into attention for egocentric activity recognition.
\newblock In {\em Proceedings of the IEEE/CVF Winter Conference on Applications of Computer Vision (WACV)}, pages 1069--1078, 2021.

\bibitem{tmm6}
Md~Moniruzzaman, Zhaozheng Yin, Zhihai He, Ruwen Qin, and Ming~C Leu.
\newblock Human action recognition by discriminative feature pooling and video segment attention model.
\newblock {\em IEEE Transactions on Multimedia}, 24:689--701, 2022.

\bibitem{narayan2014action}
Sanath Narayan, Mohan~S Kankanhalli, and Kalpathi~R Ramakrishnan.
\newblock Action and interaction recognition in first-person videos.
\newblock In {\em Proceedings of the IEEE/CVF Conference on Computer Vision and Pattern Recognition Workshops (CVPRW)}, pages 512--518, 2014.

\bibitem{vln_imgcls}
Meike Nauta, J\"org Schl\"otterer, Maurice van Keulen, and Christin Seifert.
\newblock Pip-net: Patch-based intuitive prototypes for interpretable image classification.
\newblock In {\em Proceedings of the IEEE/CVF Conference on Computer Vision and Pattern Recognition (CVPR)}, pages 2744--2753, 2023.

\bibitem{xclip}
Bolin Ni, Houwen Peng, Minghao Chen, Songyang Zhang, Gaofeng Meng, Jianlong Fu, Shiming Xiang, and Haibin Ling.
\newblock Expanding language-image pretrained models for general video recognition.
\newblock In {\em Proceedings of the European Conference on Computer Vision (ECCV)}, pages 1--18, 2022.

\bibitem{GPT-4}
OpenAI.
\newblock {GPT-4} technical report.
\newblock {\em arXiv preprint arXiv:2303.08774}, 2023.

\bibitem{gpt4v}
OpenAI.
\newblock Gpt-4v(ision) system card.
\newblock 2023.

\bibitem{st-adapter}
Junting Pan, Ziyi Lin, Xiatian Zhu, Jing Shao, and Hongsheng Li.
\newblock St-adapter: Parameter-efficient image-to-video transfer learning.
\newblock In {\em Advances in Neural Information Processing Systems (NeurIPS)}, pages 26462--26477, 2022.

\bibitem{tmm4}
Chen Pang, Xuequan Lu, and Lei Lyu.
\newblock Skeleton-based action recognition through contrasting two-stream spatial-temporal networks.
\newblock {\em IEEE Transactions on Multimedia}, 25:8699--8711, 2023.

\bibitem{pytorch}
Adam Paszke, Sam Gross, Francisco Massa, Adam Lerer, James Bradbury, Gregory Chanan, Trevor Killeen, Zeming Lin, Natalia Gimelshein, Luca Antiga, et~al.
\newblock Pytorch: An imperative style, high-performance deep learning library.
\newblock In {\em Advances in Neural Information Processing Systems (NeurIPS)}, pages 8026–--8037, 2019.

\bibitem{patrick2021keeping}
Mandela Patrick, Dylan Campbell, Yuki Asano, Ishan Misra, Florian Metze, Christoph Feichtenhofer, Andrea Vedaldi, and Jo{\~a}o~F Henriques.
\newblock Keeping your eye on the ball: Trajectory attention in video transformers.
\newblock In {\em Advances in Neural Information Processing Systems (NeurIPS)}, pages 12493--12506, 2021.

\bibitem{plizzari2022e2}
Chiara Plizzari, Mirco Planamente, Gabriele Goletto, Marco Cannici, Emanuele Gusso, Matteo Matteucci, and Barbara Caputo.
\newblock E2 (go) motion: Motion augmented event stream for egocentric action recognition.
\newblock In {\em Proceedings of the IEEE/CVF Conference on Computer Vision and Pattern Recognition (CVPR)}, pages 19935--19947, 2022.

\bibitem{EgoVLPv2_ICCV2023}
Shraman Pramanick, Yale Song, Sayan Nag, Kevin~Qinghong Lin, Hardik Shah, Mike~Zheng Shou, Rama Chellappa, and Pengchuan Zhang.
\newblock Egovlpv2: Egocentric video-language pre-training with fusion in the backbone.
\newblock In {\em Proceedings of the IEEE/CVF International Conference on Computer Vision (ICCV)}, pages 5285--5297, 2023.

\bibitem{tmm5}
Cheng Qi, Zhiyong Feng, Meng Xing, Yong Su, Jinqing Zheng, and Yiming Zhang.
\newblock Energy-based temporal summarized attentive network for zero-shot action recognition.
\newblock {\em IEEE Transactions on Multimedia}, 25:1940--1953, 2023.

\bibitem{qiu2017learning}
Zhaofan Qiu, Ting Yao, and Tao Mei.
\newblock Learning spatio-temporal representation with pseudo-3d residual networks.
\newblock In {\em Proceedings of the IEEE/CVF International Conference on Computer Vision (ICCV)}, pages 5533--5541, 2017.

\bibitem{vanila_clip}
Alec Radford, Jong~Wook Kim, Chris Hallacy, Aditya Ramesh, Gabriel Goh, Sandhini Agarwal, Girish Sastry, Amanda Askell, Pamela Mishkin, Jack Clark, Gretchen Krueger, and Ilya Sutskever.
\newblock Learning transferable visual models from natural language supervision.
\newblock In {\em Proceedings of the 38th International Conference on Machine Learning (ICML)}, pages 8748--8763, 2021.

\bibitem{ragusa2021meccano}
Francesco Ragusa, Antonino Furnari, Salvatore Livatino, and Giovanni~Maria Farinella.
\newblock The meccano dataset: Understanding human-object interactions from egocentric videos in an industrial-like domain.
\newblock In {\em Proceedings of the IEEE/CVF Winter Conference on Applications of Computer Vision (WACV)}, pages 1569--1578, 2021.

\bibitem{ren2023chatgpt}
Zhiyuan Ren, Yiyang Su, and Xiaoming Liu.
\newblock Chatgpt-powered hierarchical comparisons for image classification.
\newblock In {\em Thirty-seventh Conference on Neural Information Processing Systems}, 2023.

\bibitem{ryoo2013first}
Michael~S Ryoo and Larry Matthies.
\newblock First-person activity recognition: What are they doing to me?
\newblock In {\em Proceedings of the IEEE/CVF Conference on Computer Vision and Pattern Recognition (CVPR)}, pages 2730--2737, 2013.

\bibitem{shi2023exploring}
Yongxin Shi, Dezhi Peng, Wenhui Liao, Zening Lin, Xinhong Chen, Chongyu Liu, Yuyi Zhang, and Lianwen Jin.
\newblock Exploring ocr capabilities of gpt-4v (ision): A quantitative and in-depth evaluation.
\newblock {\em arXiv preprint arXiv:2310.16809}, 2023.

\bibitem{shu2015weakly}
Xiangbo Shu, Guo-Jun Qi, Jinhui Tang, and Jingdong Wang.
\newblock Weakly-shared deep transfer networks for heterogeneous-domain knowledge propagation.
\newblock In {\em Proceedings of the ACM International Conference on Multimedia (ACM MM)}, pages 35--44, 2015.

\bibitem{shu2015personalized}
Xiangbo Shu, Jinhui Tang, Hanjiang Lai, Luoqi Liu, and Shuicheng Yan.
\newblock Personalized age progression with aging dictionary.
\newblock In {\em Proceedings of the IEEE/CVF International Conference on Computer Vision (ICCV)}, pages 3970--3978, 2015.

\bibitem{shu2019hierarchical}
Xiangbo Shu, Jinhui Tang, Guo-Jun Qi, Wei Liu, and Jian Yang.
\newblock Hierarchical long short-term concurrent memory for human interaction recognition.
\newblock {\em IEEE Transactions on Pattern Analysis and Machine Intelligence}, 43(3):1110--1118, 2019.

\bibitem{shu2022multi}
Xiangbo Shu, Binqian Xu, Liyan Zhang, and Jinhui Tang.
\newblock Multi-granularity anchor-contrastive representation learning for semi-supervised skeleton-based action recognition.
\newblock {\em IEEE Transactions on Pattern Analysis and Machine Intelligence}, 2022.

\bibitem{shu2022expansion}
Xiangbo Shu, Jiawen Yang, Rui Yan, and Yan Song.
\newblock Expansion-squeeze-excitation fusion network for elderly activity recognition.
\newblock {\em IEEE Transactions on Circuits and Systems for Video Technology}, 32(8):5281--5292, 2022.

\bibitem{shu2021spatiotemporal}
Xiangbo Shu, Liyan Zhang, Guo-Jun Qi, Wei Liu, and Jinhui Tang.
\newblock Spatiotemporal co-attention recurrent neural networks for human-skeleton motion prediction.
\newblock {\em IEEE Transactions on Pattern Analysis and Machine Intelligence}, 44(6):3300--3315, 2021.

\bibitem{shu2020host}
Xiangbo Shu, Liyan Zhang, Yunlian Sun, and Jinhui Tang.
\newblock Host--parasite: Graph lstm-in-lstm for group activity recognition.
\newblock {\em IEEE Transactions on Neural Networks and Learning Systems}, 32(2):663--674, 2020.

\bibitem{sigurdsson2018actor}
Gunnar~A Sigurdsson, Abhinav Gupta, Cordelia Schmid, Ali Farhadi, and Karteek Alahari.
\newblock Actor and observer: Joint modeling of first and third-person videos.
\newblock In {\em proceedings of the IEEE/CVF conference on computer vision and pattern recognition (CVPR)}, pages 7396--7404, 2018.

\bibitem{Charades-ego}
Gunnar~A Sigurdsson, Abhinav Gupta, Cordelia Schmid, Ali Farhadi, and Karteek Alahari.
\newblock Charades-ego: A large-scale dataset of paired third and first person videos.
\newblock {\em arXiv preprint arXiv:1804.09626}, 2018.

\bibitem{simonyan2014two}
Karen Simonyan and Andrew Zisserman.
\newblock Two-stream convolutional networks for action recognition in videos.
\newblock In {\em Advances in Neural Information Processing Systems (NeurIPS)}, pages 568--576, 2014.

\bibitem{song2016multimodal}
Sibo Song, Vijay Chandrasekhar, Bappaditya Mandal, Liyuan Li, Joo-Hwee Lim, Giduthuri Sateesh~Babu, Phyo Phyo~San, and Ngai-Man Cheung.
\newblock Multimodal multi-stream deep learning for egocentric activity recognition.
\newblock In {\em Proceedings of the IEEE/CVF Conference on Computer Vision and Pattern Recognition Workshops (CVPRW)}, pages 24--31, 2016.

\bibitem{soomro2012ucf101}
Khurram Soomro, Amir~Roshan Zamir, and Mubarak Shah.
\newblock Ucf101: A dataset of 101 human actions classes from videos in the wild.
\newblock {\em arXiv preprint arXiv:1212.0402}, 2012.

\bibitem{stroud2020learning}
Jonathan~C Stroud, Zhichao Lu, Chen Sun, Jia Deng, Rahul Sukthankar, Cordelia Schmid, and David~A Ross.
\newblock Learning video representations from textual web supervision.
\newblock {\em arXiv preprint arXiv:2007.14937}, 2020.

\bibitem{sudhakaran2019lsta}
Swathikiran Sudhakaran, Sergio Escalera, and Oswald Lanz.
\newblock Lsta: Long short-term attention for egocentric action recognition.
\newblock In {\em Proceedings of the IEEE/CVF Conference on Computer Vision and Pattern Recognition (CVPR)}, pages 9954--9963, 2019.

\bibitem{sudhakaran2018attention}
Swathikiran Sudhakaran and Oswald Lanz.
\newblock Attention is all we need: Nailing down object-centric attention for egocentric activity recognition.
\newblock In {\em Proceedings of the British Machine Vision Conference (BMVC)}, pages 1--12, 2018.

\bibitem{tang2019coherence}
Jinhui Tang, Xiangbo Shu, Rui Yan, and Liyan Zhang.
\newblock Coherence constrained graph lstm for group activity recognition.
\newblock {\em IEEE Transactions on Pattern Analysis and Machine Intelligence}, 44(2):636--647, 2019.

\bibitem{LLAMA}
Hugo Touvron, Thibaut Lavril, Gautier Izacard, Xavier Martinet, Marie{-}Anne Lachaux, Timoth{\'{e}}e Lacroix, Baptiste Rozi{\`{e}}re, Naman Goyal, Eric Hambro, Faisal Azhar, Aur{\'{e}}lien Rodriguez, Armand Joulin, Edouard Grave, and Guillaume Lample.
\newblock Llama: Open and efficient foundation language models.
\newblock {\em arXiv preprint arXiv:2302.13971}, 2023.

\bibitem{tran2018closer}
Du~Tran, Heng Wang, Lorenzo Torresani, Jamie Ray, Yann LeCun, and Manohar Paluri.
\newblock A closer look at spatiotemporal convolutions for action recognition.
\newblock In {\em Proceedings of the IEEE/CVF Conference on Computer Vision and Pattern Recognition (CVPR)}, pages 6450--6459, 2018.

\bibitem{vaswani2017attention}
Ashish Vaswani, Noam Shazeer, Niki Parmar, Jakob Uszkoreit, Llion Jones, Aidan~N Gomez, {\L}ukasz Kaiser, and Illia Polosukhin.
\newblock Attention is all you need.
\newblock In {\em Advances in Neural Information Processing Systems (NeurIPS)}, pages 5998--6008, 2017.

\bibitem{mm22hobj}
Guangzhi Wang, Yangyang Guo, Yongkang Wong, and Mohan Kankanhalli.
\newblock Distance matters in human-object interaction detection.
\newblock In {\em Proceedings of the ACM International Conference on Multimedia (ACM MM)}, page 4546–4554, 2022.

\bibitem{mm22videosum}
Junbo Wang, Wei Wang, Zhiyong Wang, Liang Wang, Dagan Feng, and Tieniu Tan.
\newblock Stacked memory network for video summarization.
\newblock In {\em Proceedings of the ACM International Conference on Multimedia (ACM MM)}, pages 836--844, 2019.

\bibitem{wang2019hallucinating}
Lei Wang, Piotr Koniusz, and Du~Q Huynh.
\newblock Hallucinating idt descriptors and i3d optical flow features for action recognition with cnns.
\newblock In {\em Proceedings of the IEEE/CVF International Conference on Computer Vision (ICCV)}, pages 8698--8708, 2019.

\bibitem{wang2016temporal}
Limin Wang, Yuanjun Xiong, Zhe Wang, Yu~Qiao, Dahua Lin, Xiaoou Tang, and Luc Van~Gool.
\newblock Temporal segment networks: Towards good practices for deep action recognition.
\newblock In {\em Proceedings of the European Conference on Computer Vision (ECCV)}, pages 20--36, 2016.

\bibitem{actionclip}
Mengmeng Wang, Jiazheng Xing, and Yong Liu.
\newblock Actionclip: A new paradigm for video action recognition.
\newblock {\em arXiv preprint arXiv:2109.08472}, 2021.

\bibitem{tc3}
Qiang Wang, Gan Sun, Jiahua Dong, Qianqian Wang, and Zhengming Ding.
\newblock Continuous multi-view human action recognition.
\newblock {\em IEEE Transactions on Circuits and Systems for Video Technology}, 32(6):3603--3614, 2022.

\bibitem{tmm2}
Rui Wang, Jun Liu, Qiuhong Ke, Duo Peng, and Yinjie Lei.
\newblock Dear-net: Learning diversities for skeleton-based early action recognition.
\newblock {\em IEEE Transactions on Multimedia}, 25:1175--1189, 2023.

\bibitem{wang2020symbiotic2}
Xiaohan Wang, Yu~Wu, Linchao Zhu, and Yi~Yang.
\newblock Symbiotic attention with privileged information for egocentric action recognition.
\newblock In {\em Proceedings of the AAAI Conference on Artificial Intelligence (AAAI)}, volume~34, pages 12249--12256, 2020.

\bibitem{wang2021interactive}
Xiaohan Wang, Linchao Zhu, Heng Wang, and Yi~Yang.
\newblock Interactive prototype learning for egocentric action recognition.
\newblock In {\em Proceedings of the IEEE/CVF International Conference on Computer Vision (ICCV)}, pages 8168--8177, 2021.

\bibitem{wang2020symbiotic}
Xiaohan Wang, Linchao Zhu, Yu~Wu, and Yi~Yang.
\newblock Symbiotic attention for egocentric action recognition with object-centric alignment.
\newblock {\em IEEE Transactions on Pattern Analysis and Machine Intelligence}, 2020.

\bibitem{wang2022internvideo}
Yi~Wang, Kunchang Li, Yizhuo Li, Yinan He, Bingkun Huang, Zhiyu Zhao, Hongjie Zhang, Jilan Xu, Yi~Liu, Zun Wang, et~al.
\newblock Internvideo: General video foundation models via generative and discriminative learning.
\newblock {\em arXiv preprint arXiv:2212.03191}, 2022.

\bibitem{wang2017spatiotemporal}
Yunbo Wang, Mingsheng Long, Jianmin Wang, and Philip~S Yu.
\newblock Spatiotemporal pyramid network for video action recognition.
\newblock In {\em Proceedings of the IEEE Conference on Computer Vision and Pattern Recognition (CVPR)}, pages 1529--1538, 2017.

\bibitem{vln_objectdet}
Zhenyu Wang, Yali Li, Xi~Chen, Ser-Nam Lim, Antonio Torralba, Hengshuang Zhao, and Shengjin Wang.
\newblock Detecting everything in the open world: Towards universal object detection.
\newblock In {\em Proceedings of the IEEE/CVF Conference on Computer Vision and Pattern Recognition (CVPR)}, pages 11433--11443, 2023.

\bibitem{wen2023road}
Licheng Wen, Xuemeng Yang, Daocheng Fu, Xiaofeng Wang, Pinlong Cai, Xin Li, Tao Ma, Yingxuan Li, Linran Xu, Dengke Shang, et~al.
\newblock On the road with gpt-4v (ision): Early explorations of visual-language model on autonomous driving.
\newblock {\em arXiv preprint arXiv:2311.05332}, 2023.

\bibitem{wu2019long}
Chao-Yuan Wu, Christoph Feichtenhofer, Haoqi Fan, Kaiming He, Philipp Krahenbuhl, and Ross Girshick.
\newblock Long-term feature banks for detailed video understanding.
\newblock In {\em Proceedings of the IEEE/CVF Conference on Computer Vision and Pattern Recognition (CVPR)}, pages 284--293, 2019.

\bibitem{wu2022memvit}
Chao-Yuan Wu, Yanghao Li, Karttikeya Mangalam, Haoqi Fan, Bo~Xiong, Jitendra Malik, and Christoph Feichtenhofer.
\newblock Memvit: Memory-augmented multiscale vision transformer for efficient long-term video recognition.
\newblock In {\em Proceedings of the IEEE/CVF Conference on Computer Vision and Pattern Recognition (CVPR)}, pages 13587--13597, 2022.

\bibitem{cap4video}
Wenhao Wu, Haipeng Luo, Bo~Fang, Jingdong Wang, and Wanli Ouyang.
\newblock Cap4video: What can auxiliary captions do for text-video retrieval?
\newblock In {\em Proceedings of the IEEE/CVF Conference on Computer Vision and Pattern Recognition (CVPR)}, 2023.

\bibitem{text4vis}
Wenhao Wu, Zhun Sun, and Wanli Ouyang.
\newblock Revisiting classifier: Transferring vision-language models for video recognition.
\newblock In {\em Proceedings of the AAAI Conference on Artificial Intelligence (AAAI)}, pages 2847--2855, 2023.

\bibitem{text4vis_ijcv}
Wenhao Wu, Zhun Sun, Yuxin Song, Jingdong Wang, and Wanli Ouyang.
\newblock Transferring vision-language models for visual recognition: A classifier perspective.
\newblock {\em International Journal of Computer Vision}, 132(2):392–409, 2024.

\bibitem{bike}
Wenhao Wu, Xiaohan Wang, Haipeng Luo, Jingdong Wang, Yi~Yang, and Wanli Ouyang.
\newblock Bidirectional cross-modal knowledge exploration for video recognition with pre-trained vision-language models.
\newblock In {\em Proceedings of the IEEE/CVF Conference on Computer Vision and Pattern Recognition (CVPR)}, 2023.

\bibitem{GPT4Vis}
Wenhao Wu, Huanjin Yao, Mengxi Zhang, Yuxin Song, Wanli Ouyang, and Jingdong Wang.
\newblock Gpt4vis: What can gpt-4 do for zero-shot visual recognition?
\newblock 2023.

\bibitem{xie2018rethinking}
Saining Xie, Chen Sun, Jonathan Huang, Zhuowen Tu, and Kevin Murphy.
\newblock Rethinking spatiotemporal feature learning: Speed-accuracy trade-offs in video classification.
\newblock In {\em Proceedings of the European Conference on Computer Vision (ECCV)}, pages 305--321, 2018.

\bibitem{videoclip}
Hu~Xu, Gargi Ghosh, Po{-}Yao Huang, Dmytro Okhonko, Armen Aghajanyan, and et~al.
\newblock Videoclip: Contrastive pre-training for zero-shot video-text understanding.
\newblock In {\em Proceedings of the Conference on Empirical Methods in Natural Language Processing}, pages 6787--6800, 2021.

\bibitem{mm22ar}
Rui Yan, Peng Huang, Xiangbo Shu, Junhao Zhang, Yonghua Pan, and Jinhui Tang.
\newblock Look less think more: Rethinking compositional action recognition.
\newblock In {\em Proceedings of the ACM International Conference on Multimedia (ACM MM)}, page 3666–3675, 2022.

\bibitem{yan2023feature}
Rui Yan, Lingxi Xie, Xiangbo Shu, Liyan Zhang, and Jinhui Tang.
\newblock Progressive instance-aware feature learning for compositional action recognition.
\newblock {\em IEEE Transactions on Pattern Analysis and Machine Intelligence}, 45(8):10317--10330, 2019.

\bibitem{yan2022multiview}
Shen Yan, Xuehan Xiong, Anurag Arnab, Zhichao Lu, Mi~Zhang, Chen Sun, and Cordelia Schmid.
\newblock Multiview transformers for video recognition.
\newblock In {\em Proceedings of the IEEE/CVF Conference on Computer Vision and Pattern Recognition (CVPR)}, pages 3333--3343, 2022.

\bibitem{FGVP}
Lingfeng Yang, Yueze Wang, Xiang Li, Xinlong Wang, and Jian Yang.
\newblock Fine-grained visual prompting.
\newblock {\em arXiv preprint arXiv:2306.04356}, 2023.

\bibitem{aim}
Taojiannan Yang, Yi~Zhu, Yusheng Xie, Aston Zhang, Chen Chen, and Mu~Li.
\newblock Aim: Adapting image models for efficient video action recognition.
\newblock {\em arXiv preprint arXiv:2302.03024}, 2023.

\bibitem{yang2023performance}
Zhichao Yang, Zonghai Yao, Mahbuba Tasmin, Parth Vashisht, Won~Seok Jang, Beining Wang, Dan Berlowitz, and Hong Yu.
\newblock Performance of multimodal gpt-4v on usmle with image: Potential for imaging diagnostic support with explanations.
\newblock {\em medRxiv}, pages 2023--10, 2023.

\bibitem{CoCa}
Jiahui Yu, Zirui Wang, Vijay Vasudevan, Legg Yeung, Mojtaba Seyedhosseini, and Yonghui Wu.
\newblock Coca: Contrastive captioners are image-text foundation models.
\newblock {\em arXiv preprint arXiv:2205.01917}, 2022.

\bibitem{Florence}
Lu~Yuan, Dongdong Chen, Yi{-}Ling Chen, and et~al. Noel~Codella.
\newblock Florence: {A} new foundation model for computer vision.
\newblock {\em arXiv preprint arXiv:2111.11432}, 2021.

\bibitem{zhang2022object}
Chuhan Zhang, Ankush Gupta, and Andrew Zisserman.
\newblock Is an object-centric video representation beneficial for transfer?
\newblock In {\em Proceedings of the Asian Conference on Computer Vision (ACCV)}, pages 1976--1994, 2022.

\bibitem{zhang2019dynamic}
Da~Zhang, Xiyang Dai, and Yuan-Fang Wang.
\newblock Dynamic temporal pyramid network: A closer look at multi-scale modeling for activity detection.
\newblock In {\em Proceedings of the Asian Conference on Computer Vision (ACCV)}, pages 712--728, 2018.

\bibitem{mm22od}
Lu~Zhang, Yang Wang, Jiaogen Zhou, Chenbo Zhang, Yinglu Zhang, Jihong Guan, Yatao Bian, and Shuigeng Zhou.
\newblock Hierarchical few-shot object detection: Problem, benchmark and method.
\newblock In {\em Proceedings of the ACM International Conference on Multimedia (ACM MM)}, page 2002–2011, 2022.

\bibitem{lavila_2023_CVPR}
Yue Zhao, Ishan Misra, Philipp Kr\"ahenb\"uhl, and Rohit Girdhar.
\newblock Learning video representations from large language models.
\newblock In {\em Proceedings of the IEEE/CVF Conference on Computer Vision and Pattern Recognition (CVPR)}, pages 6586--6597, 2023.

\bibitem{zhou2018temporal}
Bolei Zhou, Alex Andonian, Aude Oliva, and Antonio Torralba.
\newblock Temporal relational reasoning in videos.
\newblock In {\em Proceedings of the European Conference on Computer Vision (ECCV)}, pages 803--818, 2018.

\bibitem{zhou2023exploring}
Peilin Zhou, Meng Cao, You-Liang Huang, Qichen Ye, Peiyan Zhang, Junling Liu, Yueqi Xie, Yining Hua, and Jaeboum Kim.
\newblock Exploring recommendation capabilities of gpt-4v (ision): A preliminary case study.
\newblock {\em arXiv preprint arXiv:2311.04199}, 2023.

\bibitem{zhu2023minigpt}
Deyao Zhu, Jun Chen, Xiaoqian Shen, Xiang Li, and Mohamed Elhoseiny.
\newblock Minigpt-4: Enhancing vision-language understanding with advanced large language models.
\newblock {\em arXiv preprint arXiv:2304.10592}, 2023.

\end{thebibliography}

\end{document}